\documentclass{article}

    \PassOptionsToPackage{numbers, compress}{natbib}

\usepackage[preprint]{neurips_2026}
\usepackage{graphicx}
\usepackage{booktabs}
\usepackage{xcolor}
\usepackage{multirow}
\usepackage{tabularx}
\usepackage{array}
\usepackage{makecell}
\usepackage{amssymb}

\usepackage{natbib}

\usepackage{longtable}
\usepackage{caption}

\usepackage[most]{tcolorbox}
\tcbuselibrary{breakable,listings}

\newtcblisting{promptbox}[2][]{%
  enhanced,
  breakable,
  colback=gray!5,
  colframe=black!70,
  colbacktitle=gray!25,
  boxrule=0.6pt,
  arc=1.5mm,
  left=2mm,
  right=2mm,
  top=2mm,
  bottom=1.5mm,
  title={#2},
  fonttitle=\bfseries,
  coltitle=black,
  boxed title style={
    colback=gray!25,
    colframe=black!50,
    boxrule=0.5pt,
    arc=1mm
  },
  listing only,
  verbatim,
  #1
}

\usepackage[utf8]{inputenc} 
\usepackage[T1]{fontenc}    
\usepackage{hyperref}       
\usepackage{url}            
\usepackage{booktabs}       
\usepackage{amsfonts}       
\usepackage{nicefrac}       
\usepackage{microtype}      
\usepackage{xcolor}         

\title{Towards a Virtual Neuroscientist: Autonomous Neuroimaging Analysis via Multi-Agent Collaboration}

\author{%
  Keqi Han\\
  Emory University\\
  \texttt{keqi.han@emory.edu} \\
  \And
  Songlin Zhao\\
  Lehigh University\\
  \texttt{soz223@lehigh.edu} \\
  \And
  Yao Su\\
  Worcester Polytechnic Institute\\
  \texttt{ysu6@wpi.edu} \\
  \And
  Xiang Li\\
  Massachusetts General Hospital\\
  Harvard University\\
  \texttt{xli60@mgh.harvard.edu} \\
  \And
  Yixuan Yuan\\
  Chinese University of Hong Kong\\
  \texttt{yxyuan@ee.cuhk.edu.hk} \\
  \And
  Lifang He\\
  Lehigh University\\
  \texttt{lih319@lehigh.edu} \\
  \And
  Carl Yang\thanks{Corresponding Author} \\
  Emory University\\
  \texttt{j.carlyang@emory.edu} \\
}

\begin{document}

\maketitle

\begin{abstract}
Transforming neuroimaging data into clinically actionable biomarkers is a knowledge-intensive and labor-intensive process. Standardized workflows such as fMRIPrep have improved robustness and efficiency, but they are statically configured and cannot reason about downstream objectives, deliberate over alternative strategies, or close the loop between intermediate evidence and subsequent decisions in the way a human researcher would.
This lack of closed-loop adaptation often leaves domain experts trapped in a cycle of manual trial-and-error to tune parameters and remediate pipeline failures, severely constraining the scalability of clinical biomarker development.
To bridge this gap, we introduce NEXUS, an autonomous multi-agent framework that integrates neuroimaging workflow execution with scientific-objective understanding.
Unlike conventional flat tool-calling agents, NEXUS adopts a code-centric execution paradigm where specialist agents collaboratively synthesize and optimize executable programs over composable domain-specific primitives. This design enables robust, long-horizon workflow construction that adapts dynamically to runtime observations.
Furthermore, we propose a hierarchical verification framework for autonomous quality control, integrating cohort-level metric screening with agentic visual inspection to drive evidence-grounded workflow remediation.
Experiments on ADHD-200 and ADNI demonstrate that NEXUS outperforms standard workflow-based baselines in predictive performance while exhibiting sophisticated agentic behaviors, including strategy exploration and adaptive refinement. The code is available at https://github.com/LearningKeqi/Virtual-Neuroscientist-NEXUS.
\end{abstract}

\section{Introduction}

Neuroimaging analysis plays a central role in modern neuroscience and clinical research, enabling researchers to characterize brain structure and function from modalities such as structural MRI (sMRI) and functional MRI (fMRI). Yet in practice, turning raw neuroimaging data into reliable scientific outcomes remains a knowledge-intensive and labor-intensive process, 
which often entails limited efficiency and reproducibility. To alleviate this burden, the neuroimaging community has made major progress toward standardization and automation through data conventions such as BIDS \cite{gorgolewski2016brain} and widely adopted frameworks like fMRIPrep \cite{esteban2019fmriprep}, which packages established practices into reusable workflows and has substantially improved robustness and reproducibility. However, these systems are still primarily designed as predefined workflows with bounded configurability: while they can implement well-engineered defaults and limited workflow-internal adaptation, they do not autonomously reason about a downstream scientific objective, explore alternative analytical strategies, or close the loop between intermediate evidence and subsequent decisions in the way a human researcher would during an end-to-end neuroimaging project.


Recent advances in large language models (LLMs) \cite{naveed2025comprehensive} and agentic systems \cite{gridach2025agentic} have shown remarkable potential for automating complex knowledge work and scientific discovery, raising a natural question: \textit{Can we build a virtual neuroscientist?} For neuroimaging analysis, however, realizing this vision faces several key challenges. \textit{Ecosystem heterogeneity}: neuroimaging analysis spans diverse software backends with backend-specific interfaces, assumptions, and intermediate representations, while many critical operations are highly specialized and not reliably recoverable from LLM knowledge at runtime. \textit{Long-horizon workflow complexity}: end-to-end analysis requires tightly coupled dataflow, structured control logic, parallel execution, and robust error handling, making conventional flat tool-calling brittle and unreliable. \textit{Quality-control-driven decision making}: neuroimaging quality control is not merely a reporting step, but a judgment-intensive decision point that must be tightly integrated into the workflow to support downstream decisions. \textit{Distributed expertise requirement}: the process involves distinct forms of expertise, including data understanding, image processing, quality assessment, and statistical modeling, which can overload a single monolithic agent and motivate a multi-agent design with specialized roles.

In this work, we take a step toward this goal by introducing NEXUS (Neuroimaging Execution and Understanding System), an autonomous multi-agent framework that bridges neuroimaging execution and scientific understanding through collaborative AI agents.
NEXUS addresses the above challenges through four key designs. First, to handle \textit{ecosystem heterogeneity}, we encapsulate specialized neuroimaging routines into composable domain-specific primitives, providing agents with reliable abstractions over diverse backends rather than requiring them to recover low-level domain operations from LLM knowledge. Second, to support \textit{long-horizon workflow construction}, we adopt a code-centric execution paradigm in which specialist agents synthesize executable programs over these primitives, enabling explicit dataflow, structured control logic, parallel execution, and robust error handling beyond flat tool-calling. Third, to realize \textit{quality-control-driven decision making}, we propose an agent-as-a-judge framework for closed-loop autonomous QC, integrating cohort-level metric screening with agentic visual inspection to generate actionable verdicts for dynamic workflow adaptation. Fourth, to meet the \textit{distributed expertise requirement}, NEXUS follows a centralized-planning, decentralized-execution design: a Supervisor Agent maintains global objective coherence, while domain-specific Professional Agents handle data understanding, preprocessing, quality control, and downstream analysis within focused action spaces. We further introduce just-in-time context injection, which dynamically selects task-relevant primitives for each specialist to maintain context efficiency and reduce distraction from irrelevant tools. Instead of assuming that a single fixed workflow should be applied uniformly, we formulate neuroimaging analysis as an adaptive sequential decision-making problem. Starting from raw data and a high-level scientific objective, the agent must orchestrate, refine, and ultimately deliver a workflow aligned with both the data and the end task.

We evaluate NEXUS on two widely used benchmark neuroimaging datasets, ADHD-200~\cite{adhd2012adhd} and ADNI~\cite{petersen2010alzheimer}, covering both functional and structural MRI analysis with distinct downstream objectives. The results show that NEXUS outperforms standard workflow-based baselines in predictive performance while maintaining a robust completion rate and exhibiting clear agentic behaviors, such as inter-agent coordination, strategy exploration, and adaptive refinement. Through ablation studies, we show that the architectural components are all necessary for stable long-horizon execution. Finally, we demonstrate that the proposed closed-loop QC module contributes to overall analytical outcomes and achieves moderate agreement with human QC assessments.

\vspace{-2mm}
\section{Related Work}
\vspace{-3mm}

\textbf{Neuroimaging Workflow Automation.} The neuroimaging community has made substantial progress toward standardization and automation through data standards such as BIDS~\cite{gorgolewski2016brain} and widely used software ecosystems, including fMRIPrep~\cite{esteban2019fmriprep}, FreeSurfer~\cite{fischl2012freesurfer}, BrainSuite~\cite{shattuck2002brainsuite}, Nipype~\cite{gorgolewski2011nipype}, and MRIQC~\cite{esteban2017mriqc}. These tools have improved robustness, reproducibility, and usability by packaging standard practices into reusable workflows. However, they largely rely on statically predefined processing pipelines, offer limited configurability, and treat QC as a post hoc reporting mechanism rather than a decision-making component, preventing them from reasoning about downstream scientific objectives, comparing alternative analytical strategies, or adapting execution based on intermediate evidence. Our work advances beyond this line of research by enabling autonomous, objective-aware, and feedback-driven analysis.

\textbf{LLM Agents for Scientific Workflows.} Recent work has increasingly explored LLM agents not only for general tool use, but also for scientific discovery and domain-specialized research automation. For example, ReAct~\cite{yao2023react} established a general reasoning-and-acting paradigm, while subsequent systems explored multi-agent collaboration and executable-code-based action spaces such as CodeAct~\cite{wang2024executable}. In parallel, scientific agent systems such as The AI Scientist~\cite{lu2024ai} aim to automate substantial parts of the research cycle, and domain-specific agents such as ChemCrow~\cite{bran2023chemcrow} and Biomni~\cite{huang2025biomni} demonstrate that coupling LLMs with expert tools, databases, and executable environments can substantially expand scientific task coverage. 
Concurrently, recent works such as MedRAX~\cite{fallahpour2025medrax}, NEURA~\cite{xie2026neura}, NeuroClaw~\cite{wang2026neuroclaw}, and Agentic LLMs~\cite{erdur2026agentic} explore LLM-based agents for neuroimaging workflows. While these approaches focus on integrating external tools and generating analysis workflows, they primarily focus on tool use, workflow execution, or broad assistance capabilities. In contrast, our work targets fully autonomous, end-to-end neuroimaging analysis, framing the problem as a long-horizon decision-making process and emphasizing adaptive workflow construction, dynamic refinement, objective-aware optimization, and closed-loop agentic QC.

\vspace{-2mm}
\section{Method}

\begin{figure}
    \centering
    \includegraphics[width=1\linewidth]{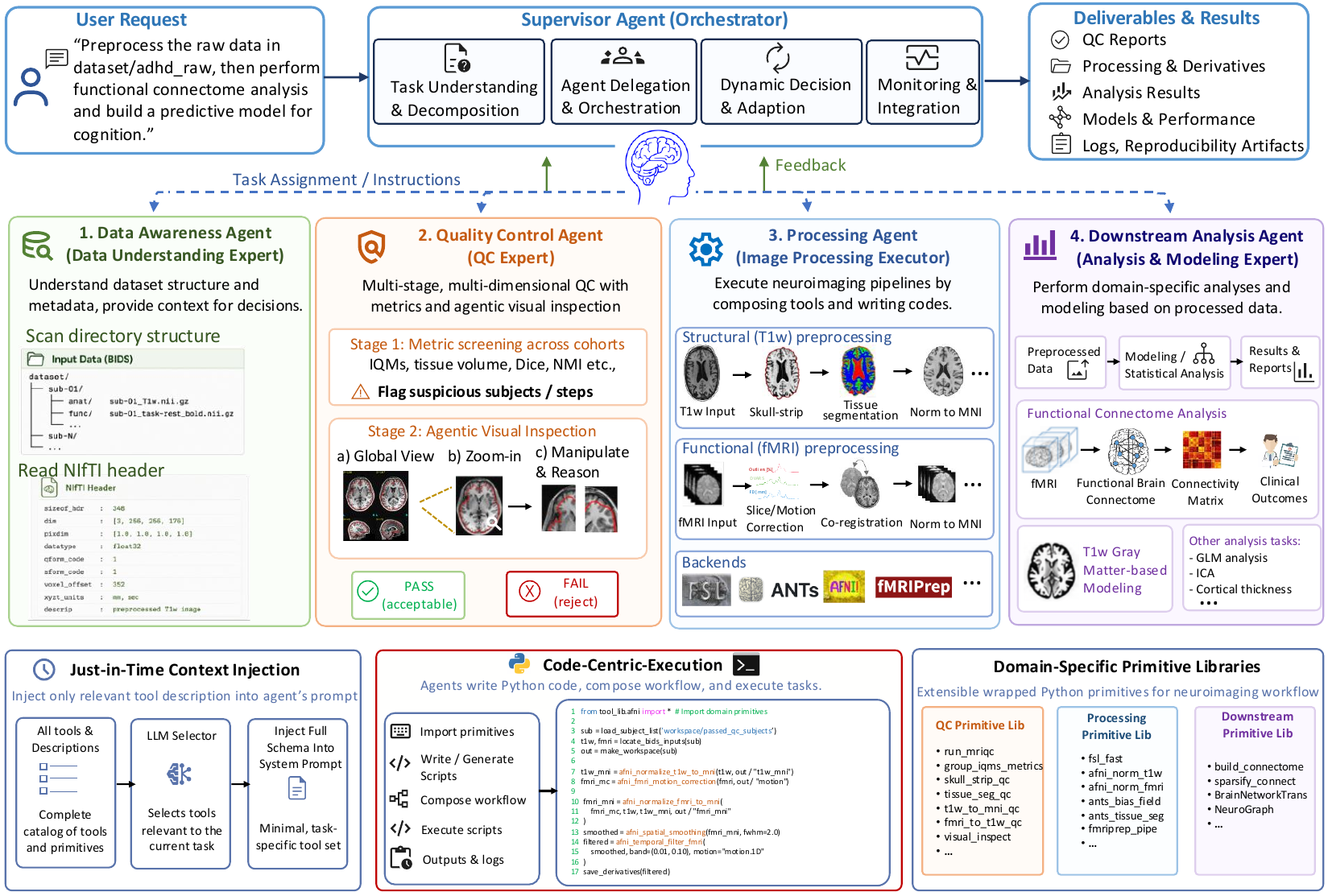}
    \caption{Overview of the NEXUS framework.}
    \label{fig:framework}
\end{figure}


\subsection{Problem Formulation and System Architecture}

\textbf{Problem Formulation.} We formalize autonomous neuroimaging analysis as a goal-oriented sequential decision-making process. Let $\mathcal{D}$ denote a raw neuroimaging dataset (e.g., T1w and fMRI in BIDS format) and $\mathcal{G}$ a high-level research objective expressed in natural language. The goal is to identify an optimal execution trace $\mathcal{T} = \{(a_1, o_1), (a_2, o_2), \dots, (a_T, o_T)\}$ that transforms $\mathcal{D}$ into a set of deliverables $\mathcal{R}$ (e.g., analytical results and validated derivatives). Here, $a_t$ denotes an action (e.g., executing a processing primitive or performing QC), and $o_t$ the observation at time $t$, capturing the updated state and execution feedback. Unlike predefined pipelines, this process is inherently adaptive: each subsequent action $a_{t+1}$ is conditioned on the cumulative history of reasoning and environmental feedback, i.e., $a_{t+1} = \pi(\mathcal{G}, \mathcal{D}, \mathcal{T}_{1:t})$, with $\pi$ denoting the agentic policy.


\textbf{Architecture.} We instantiate this formulation using a hierarchical multi-agent architecture with centralized planning and decentralized execution. As illustrated in Figure \ref{fig:framework}, the system consists of a high-level Supervisor Agent and a cohort of Domain-Specific Professional Agents: \textbf{Supervisor Agent} serves as the central orchestrator. It translates the objective $\mathcal{G}$ into an executable workflow, dispatches tasks, coordinates inter-agent collaboration, and performs dynamic decision-making based on runtime feedback; \textbf{Data Awareness Agent} extracts task-related dataset context (e.g., directory structure or metadata) to ground supervisor's planning decisions; \textbf{Quality Control Agent} operates as the critic, performing hierarchical quality assessment of neuroimaging data and issuing evidence-backed verdicts for closed-loop workflow remediation and adaptation; \textbf{Processing Agent} instantiates and executes processing plans by composing primitives (e.g., skull stripping, registration) into runnable scripts, producing derivatives with provenance; \textbf{Downstream Analysis Agent} conducts  statistical analysis or predictive modeling (e.g., connectome analysis) on QC-validated derivatives, translating processed data into objective-aligned analytical outcomes.

This architecture is designed to decouple strategic global reasoning from specialized local execution. Centralized planning ensures global objective coherence and consistent decision-making across the long-horizon neuroimaging workflow. Decentralized execution allows each specialist to operate on bounded, domain-specific contexts with well-scoped action spaces, mitigating cognitive overload and reasoning drift typically observed in monolithic agents. The architecture's modular nature affords an evolvable ecosystem where new capabilities can be added through simple expert registration.


\subsection{Code-Centric Execution with Composable Primitives}

Neuroimaging analysis with LLM agents presents several intertwined challenges: (i) The neuroimaging ecosystem is highly heterogeneous, comprising numerous software packages and specialized analytical routines, each with distinct interfaces, implicit assumptions, and incompatible intermediate representations. Naively exposing these low-level utilities to an agent exacerbates hallucination risks and unstable execution, and often leads to weak cross-step compositionality. Moreover, many critical steps are highly domain-specific and sparsely represented in LLM training data, making it unreliable for an agent to implement them at runtime. 
(ii) In conventional tool-calling paradigms, complex workflows are realized as sequences of discrete tool invocations. While sufficient for short interactions, this design becomes fragile for long-horizon scientific workflows that require tightly coupled dataflow, structured control logic (e.g., iterations over cohorts, conditional branching), parallel execution, and robust error handling. As task complexity scales, these limitations manifest as brittle compositionality and reduced execution reliability. To address these challenges, inspired by CodeAct~\cite{wang2024executable}, we adopt a code-centric execution paradigm built upon composable domain-specific primitives. Instead of exposing raw tools as flat callable actions, we encapsulate heterogeneous functionalities into standardized computational primitives and empower agents to synthesize flexible programs that orchestrate them.

\subsubsection{Domain-Specific Primitives}
We define a primitive as an atomic, composable operation over neuroimaging artifacts. Formally, each primitive $p \in \mathcal{P}$ is a Python callable, specified by a \emph{card} with (i) a name, module path, and short description, and (ii) a detailed I/O schema describing required inputs, outputs, and side effects, wrapping established software (e.g., neuroimaging processing softwares, QC softwares) or custom utilities, and returning structured outputs:
\begin{equation}
p: (x, \theta) \rightarrow (x', r),
\end{equation}
where $x$ are input artifacts, $\theta$ are parameters, $x'$ are output artifacts, and $r$ is the returned result (e.g., QC verdicts, predictive performance). Primitives are organized into three families aligned with Professional Agents:
(a) Processing primitives implement image transformations and preprocessing steps across sMRI and fMRI;
(b) QC primitives compute quantitative metrics and generate visualization elements for inspection;
(c) Analysis primitives implement downstream feature construction and modeling components (e.g., ROI time-series extraction, connectome construction, predictive models).
These primitives provide a standardized abstraction layer over heterogeneous neuroimaging backends and specialized analytical routines, shielding agents from low-level interface differences and reducing the need to synthesize fragile domain-specific implementations at runtime.

\vspace{-1mm}

\subsubsection{Just-in-Time Context Injection}

Naively injecting all primitive descriptions and schemas into the prompt can lead to context inflation and degraded reasoning, particularly when the number of available primitives is large. We therefore introduce Just-in-Time (JIT) Context Injection: before a Professional Agent enters its execution loop, it performs retrieval over primitive library and injects only the \emph{minimal sufficient} context into its system prompt.

Each Professional Agent owns a primitive library $\mathcal{P}$, where each primitive $p \in \mathcal{P}$ is associated with a name $n_p$ and a short description $d_p$, in addition to a full schema (parameters, preconditions, outputs). We build a compact index consisting only of $\{(n_p, d_p)\}_{p\in\mathcal{P}}$. Given a Supervisor-issued instruction $I$, a selector produces a task-relevant subset of primitives:
\begin{equation}
\mathcal{P}_I = \texttt{SelectTools}\!\left(I,\{(n_p, d_p)\}_{p\in\mathcal{P}}\right),
\end{equation}
where $\texttt{SelectTools}(\cdot)$ denotes a single LLM invocation prompted to select a task-relevant subset of primitives from the compact index, and $\mathcal{P}_I \subseteq \mathcal{P}$ (typically $|\mathcal{P}_I| \ll |\mathcal{P}|$). Only primitives in $\mathcal{P}_I$ have their full schemas injected into the agent prompt. This just-in-time injection (i) reduces token load, (ii) mitigates distraction from irrelevant affordances, and (iii) improves modular extensibility since adding new primitives does not linearly inflate every prompt.

\vspace{-1mm}
\subsubsection{Code-Centric Synthesis}

Instead of emitting sequential, isolated tool calls, the core action of the Professional Agents is the synthesis and execution of programs. Given a supervisor-issued instruction $I$ and the professional's own cumulative history $\mathcal{T}$, a professional agent can produce an executable program $\mathcal{C}$ that composes primitives with explicit control flow:
\begin{equation}
\mathcal{C} = \textsc{Synthesize}(I, \mathcal{P}_I, \mathcal{T}),
\end{equation}
where $\mathcal{P}_I$ is the task-related primitive set selected via just-in-time context injection. The program $\mathcal{C}$ is executed in the runtime environment, and its outputs (artifacts, logs, metrics) are incorporated back into $\mathcal{T}$ for subsequent reasoning. Compared with flat tool invocation, this "code-as-act" design enables expressive control logic, explicit data provenance, and scalable composition, supporting flexible and reproducible long-horizon neuroimaging workflows.

In addition to code synthesis over domain-specific primitives, professional agents are also equipped with a minimal set of generic execution utilities (e.g., file I/O and controlled shell execution) to interface with the runtime environment and manage artifacts.

\vspace{-1mm}
\subsection{Agent-as-a-Judge for Closed-loop Autonomous QC}

A central bottleneck in neuroimaging analysis is that \emph{QC is both indispensable and difficult to automate}. Existing pipelines can generate QC reports, but converting reports into actionable decisions (e.g., exclusion, parameter adjustment, or alternative backends) remains largely manual. We formalize QC as an evidence-grounded judgment problem and instantiate the Quality Control Agent as an \emph{Agent-as-a-Judge} operating in a closed loop with the Supervisor and Processing Agents.

\vspace{-1mm}

\subsubsection{Hierarchical QC as Structured Judgment}


Let $x$ denote a neuroimaging artifact (raw data or derivative). The QC Agent adaptively constructs a set of quantitative image quality metrics $\phi(x)$ and corresponding visualization artifacts $v(x)$ conditioned on the processing stage and the provenance of $x$. The QC problem is to produce a binary verdict $y \in \{\texttt{PASS}, \texttt{FAIL}\}$ and an evidence bundle $e$ that supports the decision:
\begin{equation}
(y, e) = \mathcal{J}(x, \phi(x), v(x)),
\end{equation}
where $\mathcal{J}$ is the structured judgment conducted by the QC Agent. The verdict $y$ and evidence $e$ are fed back to the Supervisor, which conditions subsequent planning and processing strategies on QC outcomes, forming a closed-loop adaptive workflow.

Concretely, QC agent adopts a two-stage hierarchical strategy. First, a metric-based screening stage performs cohort-level outlier detection over $\{\phi(x_i)\}_{i=1}^N$, identifying a subset $\mathcal{S} \subset \{1,\dots,N\}$ of potentially problematic subjects, which reduces the search space and enables scalable automation. Second, for $i \in \mathcal{S}$, the agent further performs fine-grained visual inspection to determine the final verdict. This hierarchy mirrors expert practice.


\vspace{-1mm}

\subsubsection{Agentic Visual Inspection}

To mimic the specialized "eye" of a neuroscientist, we empower the QC Agent with Agentic Visual Inspection. When a visual check is needed, the agent invokes a Vision-Language Model (VLM) tasked with a multi-turn diagnostic protocol. The visualization $v(x)$ constructed by QC agent is passed to the VLM configured with:
(i) explicit scoring criteria describing acceptable and unacceptable patterns;
(ii) curated few-shot exemplars illustrating typical success and failure modes; and
(iii) access to a controlled programming tool enabling iterative region-of-interest cropping and zoom-in inspection. Instead of producing a single-shot judgment, the VLM engages in iterative analysis: it may proactively zoom into fine-grained details with the given programming tool, inspect boundary regions, and accumulate intermediate observations before emitting a final verdict. This \emph{program-augmented visual reasoning} mitigates coarse global judgments and improves sensitivity to details.

\vspace{-2mm}
\vspace{-1mm}

\section{Experiments}

\vspace{-1mm}

To evaluate the efficacy and autonomy of NEXUS, we design extensive experiments to answer three primary research questions: \textbf{(RQ1)} Can the system transform raw neuroimaging data and a high-level scientific objective into well-adapted analytical workflows and outcome-oriented solutions through autonomous exploration, iterative refinement, and dynamic decision-making? \textbf{(RQ2)} Are the architectural design choices of the system necessary for stable and efficient long-horizon execution?  \textbf{(RQ3)} Does the proposed QC module contribute meaningfully to overall analytical outcomes, and to what extent do its judgments align with human assessment?

\vspace{-2mm}

\subsection{Experimental Setup}

\vspace{-1mm}

\textbf{Datasets and Tasks.} We evaluate NEXUS on two widely used public neuroimaging datasets that represent distinct modalities and downstream objectives: ADHD-200~\cite{adhd2012adhd} (fMRI, $N=200$; 100 ADHD, 100 Healthy Controls) and ADNI~\cite{petersen2010alzheimer} (sMRI, $N=200$; 100 Alzheimer's Disease, 100 Cognitively Normal). For each dataset, the agent is provided with a natural-language objective specifying the prediction target and required deliverables, and is tasked with developing a complete neuroimaging analysis workflow from the raw training data. Concretely, this requires the agent to inspect the dataset, design and orchestrate an appropriate preprocessing and quality-control strategy, carry out downstream analysis and model training, and adaptively refine the workflow based on intermediate results and execution feedback. The final deliverable consists of the developed preprocessing pipeline, and the trained model artifacts along with the corresponding inference procedure, such that the resulting analysis package can be readily reusable and generalized to unseen held-out subjects for final evaluation. The two datasets instantiate this general setting with different modality-specific objectives. For ADHD-200, the agent must derive an end-to-end workflow from raw fMRI data for functional connectome-based diagnosis prediction. For ADNI, the corresponding objective is to derive such a workflow from raw T1-weighted MRI for structural diagnosis prediction based on voxel-wise gray matter features. Detailed task prompts are provided in Appendix \ref{app:task_prompts}.

\textbf{Baselines.} We compare NEXUS with standard predefined workflow-based neuroimaging baselines. These baselines represent the dominant non-agentic paradigm in current practice: fixed preprocessing pipelines instantiated with standard neuroimaging software stacks, followed by automated downstream model selection on the training set. Concretely, the evaluated preprocessing backbones include fMRIPrep~\cite{esteban2019fmriprep}, AFNI~\cite{cox1996afni}, FSL~\cite{jenkinson2012fsl}, SPM12~\cite{ashburner2014spm12} and ANTs~\cite{avants2009advanced}. The first three tools support both structural and functional MRI and are thus applied to both datasets. Since SPM12 is primarily designed for fMRI, we evaluate it only on ADHD, whereas ANTs, tailored for sMRI, is evaluated only on ADNI. For all baselines, preprocessing follows standard workflow configurations commonly used for each software package.
Downstream analysis is implemented as an automated cross-validation procedure on the training set, searching over model classes, hyperparameters, and task-specific settings. After model selection, the chosen model is retrained on the full training set and evaluated once on the held-out test set. Detailed baseline workflow specifications can be found in Appendix \ref{ablation:baseline_specifications}.

\textbf{Evaluation Protocol.} For each dataset, we use a stratified 4:1 split of the 200 subjects: 160 for workflow development, cross-validation, and model selection, and 40 held out for final evaluation. NEXUS and all baselines are run five times; we report mean held-out AUC, F1, and Accuracy. For NEXUS, we also report execution statistics. Agents use \texttt{gpt-5.2} with default settings, and visual QC uses \texttt{gemini-3-flash-preview}.

\vspace{-1mm}

\subsection{End-to-End Autonomous Neuroimaging Analysis (RQ1)}

\textbf{Downstream Predictive Performance.} Table \ref{tab:main_results} compares the held-out test performance of NEXUS against predefined workflow-based baselines on ADHD-200 and ADNI. Across both datasets, NEXUS achieves the best overall performance, indicating that adaptive, closed-loop workflow construction can yield stronger final predictive performance than predefined workflow-based alternatives. The key distinction lies in how preprocessing, quality control, and downstream modeling are coupled. The predefined baselines are automated but static, whereas NEXUS can dynamically develop and adapt its analysis strategy based on dataset characteristics, task requirements, QC outcomes, and intermediate execution feedback.

\begin{table*}[t]
\centering
\caption{End-to-end downstream prediction performance (\%) on held-out test sets. DS denotes the automated downstream analysis used on top of each predefined preprocessing backbone.}
\label{tab:main_results}
\renewcommand{\arraystretch}{1.15}
\setlength{\tabcolsep}{4.5pt}
\resizebox{\textwidth}{!}{%
\begin{tabular}{lccc|lccc}
\toprule
& \multicolumn{3}{c|}{\textbf{ADHD}} & & \multicolumn{3}{c}{\textbf{ADNI}} \\
\cline{2-4} \cline{6-8}
\textbf{Method} & \textbf{AUC} & \textbf{F1} & \textbf{ACC} & \textbf{Method} & \textbf{AUC} & \textbf{F1} & \textbf{ACC} \\
\midrule
fMRIPrep + DS & 61.30 $\pm$ 2.75 & 61.89 $\pm$ 6.32 & 54.50 $\pm$ 3.67 & fMRIPrep + DS & 90.60 $\pm$ 1.62 & 81.06 $\pm$ 2.40 & 83.00 $\pm$ 1.87 \\
AFNI + DS     & 62.60 $\pm$ 5.76 & 58.09 $\pm$ 6.48  & 57.00 $\pm$ 3.67 & AFNI + DS     & 89.30 $\pm$ 0.46 & 79.56 $\pm$ 0.89 & 82.00 $\pm$ 1.00 \\
FSL + DS      & 55.40 $\pm$ 6.50 & 61.40 $\pm$ 4.80  & 55.00 $\pm$ 5.20 & FSL + DS      & 80.90 $\pm$ 0.90 & 69.50 $\pm$ 3.90 & 71.00 $\pm$ 3.70 \\
SPM + DS      & 57.20 $\pm$ 10.30 & 64.50 $\pm$ 3.40 & 54.00 $\pm$ 7.40 & ANTs + DS     & 89.30 $\pm$ 0.50 & 75.10 $\pm$ 3.90 & 76.50 $\pm$ 3.40 \\
\midrule
\textbf{NEXUS (Ours)} & \textbf{65.55 $\pm$ 3.66} & \textbf{65.70 $\pm$ 3.01} & \textbf{65.40 $\pm$ 4.35} & \textbf{NEXUS (Ours)} & \textbf{93.00 $\pm$ 1.94} & \textbf{82.83 $\pm$ 2.46} & \textbf{84.50 $\pm$ 2.09} \\
\bottomrule
\end{tabular}%
}
\end{table*}

\begin{table*}[ht]
\centering
\small
\setlength{\tabcolsep}{4pt}
\renewcommand{\arraystretch}{1.08}

\caption{Summary statistics across five independent runs of NEXUS. Values are reported as mean (range) across runs. Completion denotes producing all required deliverables; adaptive modeling refinement denotes downstream-analysis iterations triggered by runtime observations such as validation performance, feature instability, or overfitting risk.}
\label{tab:adhd_adni_summary}

\begin{tabularx}{\textwidth}{
>{\raggedright\arraybackslash}m{0.27\textwidth}
>{\raggedright\arraybackslash}m{0.34\textwidth}
>{\raggedright\arraybackslash}m{0.34\textwidth}}
\toprule
\textbf{Statistic} & \textbf{ADHD} & \textbf{ADNI} \\
\midrule

\multicolumn{3}{l}{\textbf{Global Execution Footprint}} \\

Completion Rate
& 5/5
& 5/5 \\

Runtime
& 54.4 h (27.8--70.9)
& 38.0 h (23.7--83.2) \\

API Cost
& \$9.01 (6.77--11.94)
& \$6.46 (3.61--10.68) \\

\# inter-agent interactions
& 25 (18--32)
& 26 (20--30) \\

\# agent actions
& 158 (132--200)
& 126 (82--159) \\

\# executable Python scripts
& 42 (27--64)
& 29 (16--48) \\

\midrule
\multicolumn{3}{l}{\textbf{Exploration and Adaptation}} \\

\# candidate preprocessing pipelines explored
& 2 (1--3)
& 2 (2--3) \\

\# adaptive modeling refinement rounds
& 3 (2--5)
& 2 (1--4) \\

\# subjects passing QC
& 130 (126--135)
& 148 (140--154) \\

QC checks conducted
& Raw T1w/BOLD MRIQC; fMRI$\to$T1w coregistration; fMRI$\to$MNI QC
& Raw T1w MRIQC; skull-stripping; tissue-segmentation; T1w$\to$MNI QC \\

Delivered preprocessing pipeline
& AFNI-based (2/5), fMRIPrep-based (3/5)
& ANTs-based (3/5), ANTs+FSL Hybrid (2/5) \\

Delivered downstream modeling
& Site-aware connectome classification; AAL90/Schaefer200/HCP360 + Pearson/Fisher-z + regularized LR or BrainNetTransformer
& Voxel-wise VBM classification; univariate feature selection + LR/ElasticNet \\
\bottomrule
\end{tabularx}
\end{table*}

\vspace{-1mm}

\textbf{Agentic Dynamics and Execution Statistics.}   
Table \ref{tab:adhd_adni_summary} shows that NEXUS is not merely executing a fixed automated recipe, but actively developing and refining workflows through multi-agent coordination and runtime adaptation. Across both datasets, it achieves a 100\% completion rate over all independent runs, while exploring multiple candidate preprocessing pipelines, conducting stage-specific QC, and performing several rounds of downstream refinement before issuing a final deliverable. This pattern indicates a context-aware planning and dynamic adjustment process rather than one-shot execution, which explains why NEXUS consistently outperforms predefined workflow-based baselines. The delivered workflows vary across datasets and runs---e.g., AFNI- versus fMRIPrep-based preprocessing with different connectome modeling configurations on ADHD-200, and pure ANTs versus ANTs+FSL hybrid preprocessing with different voxel-wise VBM modeling strategies on ADNI---showing that the system does not commit to a single hard-coded software stack or downstream recipe, but instead makes data- and task-dependent decisions. The substantial number of supervisor-subagent interactions and executable scripts further shows that successful execution relies on sustained multi-agent coordination and code-centric orchestration, which naturally accommodates the complex dataflows and dependencies inherent in neuroimaging analysis. Detailed running trajectory can be found in Appendix \ref{app:example_trajectory}.

\vspace{-1mm}

\subsection{Architectural Ablations (RQ2)}

\begin{figure}[htbp]
  \centering
  \includegraphics[width=1\textwidth]{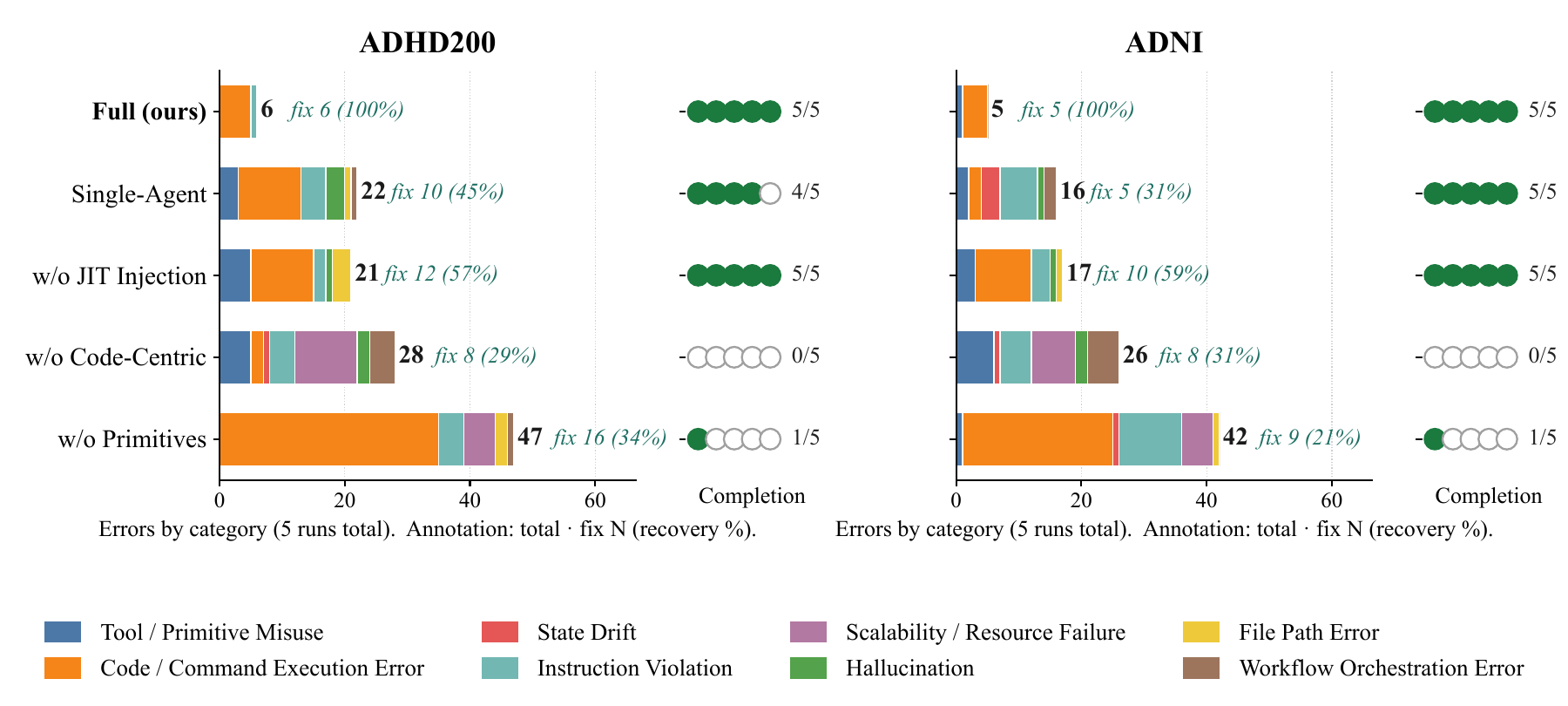}
  \caption{Ablation study results. Stacked bars show total execution errors across five independent runs. Annotations report total error count, number of autonomously recovered errors, and recovery rate. Filled and open circles indicate completed and failed runs.}
  \label{fig:ablation study}
\end{figure}

\vspace{-1mm}

We next evaluate whether the main architectural designs of NEXUS are necessary for stable execution. We run the ablation study on subsampled 20-subject subsets of ADHD-200 and ADNI. The task formulation remains the same as in the full end-to-end setting, but the goal here is not to compare downstream predictive performance. Instead, we focus on execution stability, completion and failure characteristics. Specifically, we compare the following variants: (1) \textbf{Single-Agent}, which collapses all roles into a single monolithic agent handling the whole workflow of neuroimaging analysis; (2) \textbf{w/o JIT Context Injection}, which removes task-adaptive primitive selection and injects the full primitive library into each sub-agent prompt; (3) \textbf{w/o Code-Centric Execution}, which replaces executable program synthesis with direct, flat tool calling; and (4) \textbf{w/o Primitive Abstraction}, which removes the encapsulated domain-specific primitives and requires agents to rely on low-level commands or self-written implementation logic. All variants are evaluated under the same execution budget, with a maximum of 500 ReAct recursion (observe $\rightarrow$ act $\rightarrow$ ...) limits and 12 hours of runtime.

As illustrated in Figure \ref{fig:ablation study}, the full system NEXUS consistently achieves a 100\% completion rate with fewest execution errors and uniquely maintains a 100\% autonomous error recovery rate. In contrast, degrading the architecture exposes distinct failure modes: 
Both the Single-Agent and w/o JIT Injection variants highlight the vulnerability of LLMs to cognitive and contextual overload. 
Forcing a monolithic agent to handle both global planning and local execution overwhelms its reasoning capacity, leading to State Drift, Hallucination, and sharply degraded error recovery (dropping to 31\% on ADNI). Similarly, omitting JIT injection bloats the prompt with irrelevant tool schemas, directly inducing elevated Tool/Primitive Misuse and Instruction Violations.
The w/o Code-Centric variant fails catastrophically (0/5 completion on both datasets). It suffers from severe Scalability/Resource Failure and Workflow Orchestration Errors, demonstrating the necessity of code-centric paradigm for long-horizon scientific workflows that require robust control flow and efficient parallel execution. Removing primitive abstraction causes one of the most severe performance drops. Without pre-encapsulated domain-specific operations, the agents must directly generate low-level commands and implementation details for a highly heterogeneous and specialized neuroimaging ecosystem. This substantially increases command misuse, artifact mismatch, and malformed execution scripts. Detailed ablation records and error taxonomy can be found in Appendix \ref{app:ablation_study}.

\vspace{-2mm}

\subsection{Evaluation of Closed-Loop Autonomous Quality Control (RQ3)}

\vspace{-2mm}
\begin{figure}[htbp]
  \centering
  \includegraphics[width=1\textwidth]{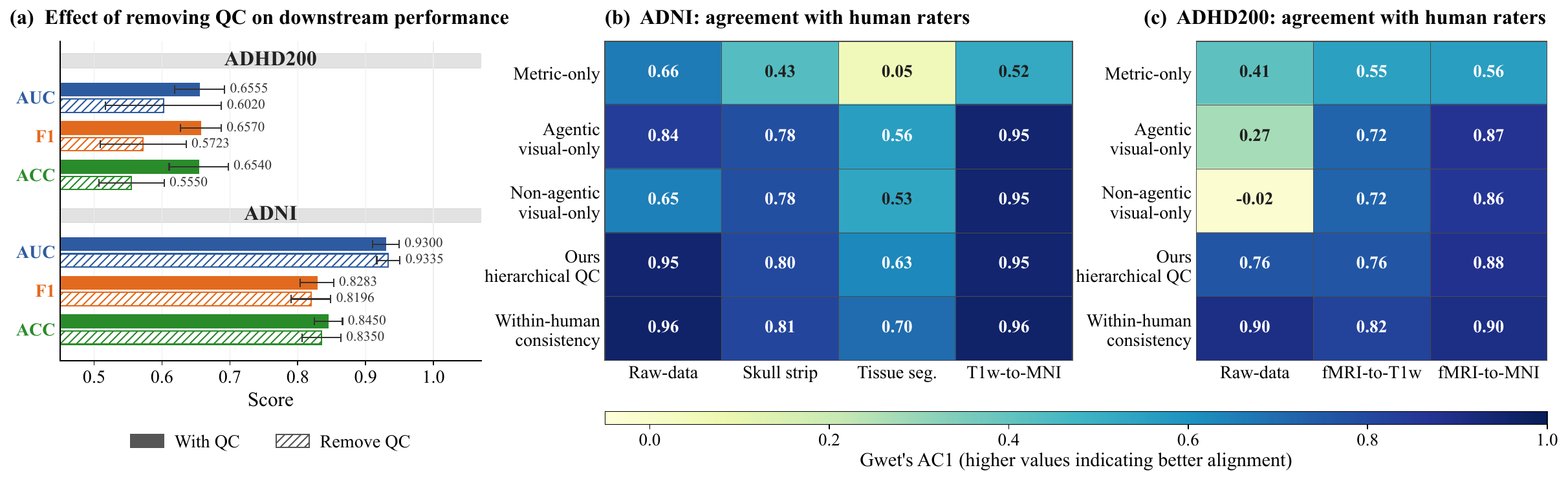}
  \caption{Evaluation of the closed loop autonomous QC module.}
  \label{fig:abaltion study}
\end{figure}

\vspace{-1mm}

\textbf{Impact of QC Filtering on Downstream Performance.} As shown in Figure \ref{fig:abaltion study}(a), removing the QC filtering stage (i.e., keeping all subjects into downstream modeling) leads to a significant performance degradation on the ADHD-200 dataset. On the ADNI dataset, we observe a more modest decrease in F1 and Accuracy. This divergence is consistent with the underlying data characteristics: ADHD-200 consists of noisier fMRI data, making downstream performance more sensitive to quality variation, whereas ADNI consists of generally higher-quality sMRI data, with fewer subjects excluded by QC. Overall, these results demonstrate that the proposed closed-loop QC effectively intercepts artifacts and poor-quality features, preventing them from confounding downstream analysis and improving the robustness of the final scientific deliverables.

\vspace{-1mm}

\textbf{Alignment with Human Assessments.} Figure \ref{fig:abaltion study}(b) and (c) reports the agreement between each QC variant and human raters across multiple checkpoints, measured by Gwet’s AC1~\cite{gwet2008computing}, an inter-rater agreement coefficient suitable for highly imbalanced settings, with higher values indicating better alignment. Each reported value is averaged over comparisons to the three human raters.
The proposed hierarchical QC achieves the strongest agreement with human judgments across all checkpoints, indicating that metric-based screening followed by visual inspection better captures both quantitative abnormalities and qualitative anatomical cues. Agentic visual QC also generally aligns better with human raters than non-agentic visual QC, indicating the value of iterative visual exploration for specialized neuroimaging judgment. Detailed QC evaluation settings and human assessments can be found in Appendix \ref{app:qc_evaluation}.

\vspace{-1.5mm}
\vspace{-1mm}

\section{Conclusion}

\vspace{-2mm}
In this work, we introduced NEXUS, a novel multi-agent system that advances neuroimaging analysis from statically predefined pipelines to autonomous, adaptive, and closed-loop scientific workflows. To overcome the high heterogeneity and long-horizon complexity inherent in neuroimaging, NEXUS leverages a code-centric execution paradigm over composable domain-specific primitives, effectively grounded by just-in-time context injection. Furthermore, we addressed the critical bottleneck of manual quality control by proposing a hierarchical, agent-as-a-judge framework. By integrating cohort-level quantitative metric screening with program-augmented agentic visual inspection, NEXUS successfully translates quality assessment into evidence-driven workflow remediation. Experiments on ADHD-200 and ADNI show that NEXUS outperforms predefined workflow-based baselines, exhibits clear agentic dynamics and relies critically on its architectural components for stable long-horizon execution. We further show that closed-loop QC both contributes to downstream outcomes and aligns well with human assessment. 
More broadly, these results highlight the promise of agentic neuroimaging analysis. Given its modular and extensible design, NEXUS provides a foundation for future systems supporting broader neuroimaging modalities, richer primitive libraries, and more diverse downstream tasks.


\newpage

\bibliographystyle{unsrtnat}
\bibliography{reference}

@article{gorgolewski2016brain,
  title={The brain imaging data structure, a format for organizing and describing outputs of neuroimaging experiments},
  author={Gorgolewski, Krzysztof J and Auer, Tibor and Calhoun, Vince D and Craddock, R Cameron and Das, Samir and Duff, Eugene P and Flandin, Guillaume and Ghosh, Satrajit S and Glatard, Tristan and Halchenko, Yaroslav O and others},
  journal={Scientific data},
  volume={3},
  number={1},
  pages={1--9},
  year={2016},
  publisher={Nature Publishing Group}
}

@article{esteban2019fmriprep,
  title={fMRIPrep: a robust preprocessing pipeline for functional MRI},
  author={Esteban, Oscar and Markiewicz, Christopher J and Blair, Ross W and Moodie, Craig A and Isik, A Ilkay and Erramuzpe, Asier and Kent, James D and Goncalves, Mathias and DuPre, Elizabeth and Snyder, Madeleine and others},
  journal={Nature methods},
  volume={16},
  number={1},
  pages={111--116},
  year={2019},
  publisher={Nature Publishing Group US New York}
}

@article{naveed2025comprehensive,
  title={A comprehensive overview of large language models},
  author={Naveed, Humza and Khan, Asad Ullah and Qiu, Shi and Saqib, Muhammad and Anwar, Saeed and Usman, Muhammad and Akhtar, Naveed and Barnes, Nick and Mian, Ajmal},
  journal={ACM Transactions on Intelligent Systems and Technology},
  volume={16},
  number={5},
  pages={1--72},
  year={2025},
  publisher={ACM New York, NY}
}

@article{gridach2025agentic,
  title={Agentic AI for Scientific Discovery: A Survey of Progress, Challenges, and Future Directions},
  author={Gridach, Mourad and Nanavati, Jay and Abidine, Khaldoun Zine El and Mendes, Lenon and Mack, Christina},
  journal={arXiv preprint arXiv:2503.08979},
  year={2025}
}

@article{fischl2012freesurfer,
  title={FreeSurfer},
  author={Fischl, Bruce},
  journal={Neuroimage},
  volume={62},
  number={2},
  pages={774--781},
  year={2012},
  publisher={Elsevier}
}

@article{shattuck2002brainsuite,
  title={BrainSuite: an automated cortical surface identification tool},
  author={Shattuck, David W and Leahy, Richard M},
  journal={Medical image analysis},
  volume={6},
  number={2},
  pages={129--142},
  year={2002},
  publisher={Elsevier}
}

@article{gorgolewski2011nipype,
  title={Nipype: a flexible, lightweight and extensible neuroimaging data processing framework in python},
  author={Gorgolewski, Krzysztof and Burns, Christopher D and Madison, Cindee and Clark, Dav and Halchenko, Yaroslav O and Waskom, Michael L and Ghosh, Satrajit S},
  journal={Frontiers in neuroinformatics},
  volume={5},
  pages={13},
  year={2011},
  publisher={Frontiers Research Foundation}
}

@article{esteban2017mriqc,
  title={MRIQC: Advancing the automatic prediction of image quality in MRI from unseen sites},
  author={Esteban, Oscar and Birman, Daniel and Schaer, Marie and Koyejo, Oluwasanmi O and Poldrack, Russell A and Gorgolewski, Krzysztof J},
  journal={PloS one},
  volume={12},
  number={9},
  pages={e0184661},
  year={2017},
  publisher={Public Library of Science San Francisco, CA USA}
}

@inproceedings{yao2023react,
  title={ReAct: Synergizing Reasoning and Acting in Language Models},
  author={Yao, Shunyu and Zhao, Jeffrey and Yu, Dian and Du, Nan and Shafran, Izhak and Narasimhan, Karthik and Cao, Yuan},
  booktitle={International Conference on Learning Representations (ICLR)},
  year={2023}
}

@inproceedings{wang2024executable,
  title={Executable code actions elicit better llm agents},
  author={Wang, Xingyao and Chen, Yangyi and Yuan, Lifan and Zhang, Yizhe and Li, Yunzhu and Peng, Hao and Ji, Heng},
  booktitle={Forty-first International Conference on Machine Learning},
  year={2024}
}

@article{lu2024ai,
  title={The ai scientist: Towards fully automated open-ended scientific discovery},
  author={Lu, Chris and Lu, Cong and Lange, Robert Tjarko and Foerster, Jakob and Clune, Jeff and Ha, David},
  journal={arXiv preprint arXiv:2408.06292},
  year={2024}
}

@article{bran2023chemcrow,
  title={Chemcrow: Augmenting large-language models with chemistry tools},
  author={Bran, Andres M and Cox, Sam and Schilter, Oliver and Baldassari, Carlo and White, Andrew D and Schwaller, Philippe},
  journal={arXiv preprint arXiv:2304.05376},
  year={2023}
}

@article{huang2025biomni,
  title={Biomni: A general-purpose biomedical ai agent},
  author={Huang, Kexin and Zhang, Serena and Wang, Hanchen and Qu, Yuanhao and Lu, Yingzhou and Roohani, Yusuf and Li, Ryan and Qiu, Lin and Li, Gavin and Zhang, Junze and others},
  journal={biorxiv},
  year={2025}
}

@inproceedings{fallahpour2025medrax,
  title={MedRAX: Medical Reasoning Agent for Chest X-ray},
  author={Fallahpour, Adibvafa and Ma, Jun and Munim, Alif and Lyu, Hongwei and Wang, Bo},
  booktitle={International Conference on Machine Learning},
  pages={15661--15676},
  year={2025},
  organization={PMLR}
}

@article{xie2026neura,
  title={NEURA: An agentic system for autonomous neuroimaging workflows},
  author={Xie, Jun and Wang, Jing and Wu, Xiumei and Liu, Xinyuan and Mi, Yiqi and Liu, Qinjin and Xu, Tong and Liu, Chen and Chen, Huafu and Guo, Jing},
  journal={bioRxiv},
  pages={2026--04},
  year={2026},
  publisher={Cold Spring Harbor Laboratory}
}

@article{wang2026neuroclaw,
  title={NeuroClaw Technical Report},
  author={Wang, Cheng and He, Zhibin and Peng, Zhihao and Liu, Shengyuan and Hu, Yufan and Sun, Lichao and Li, Xiang and Yuan, Yixuan},
  journal={arXiv preprint arXiv:2604.24696},
  year={2026}
}

@article{erdur2026agentic,
  title={Agentic Large Language Models for Training-Free Neuro-Radiological Image Analysis},
  author={Erdur, Ayhan Can and Scholz, Daniel and Pan, Jiazhen and Wiestler, Benedikt and Rueckert, Daniel and Peeken, Jan C},
  journal={arXiv preprint arXiv:2604.16729},
  year={2026}
}

@article{adhd2012adhd,
  title={The ADHD-200 consortium: a model to advance the translational potential of neuroimaging in clinical neuroscience},
  author={ADHD-200 consortium},
  journal={Frontiers in systems neuroscience},
  volume={6},
  pages={62},
  year={2012},
  publisher={Frontiers Research Foundation}
}

@article{petersen2010alzheimer,
  title={Alzheimer's disease Neuroimaging Initiative (ADNI) clinical characterization},
  author={Petersen, Ronald Carl and Aisen, Paul S and Beckett, Laurel A and Donohue, Michael C and Gamst, Anthony Collins and Harvey, Danielle J and Jack Jr, Clifford R and Jagust, William J and Shaw, Leslie M and Toga, Arthur W and others},
  journal={Neurology},
  volume={74},
  number={3},
  pages={201--209},
  year={2010},
  publisher={Lippincott Williams \& Wilkins}
}

@article{cox1996afni,
  title={AFNI: software for analysis and visualization of functional magnetic resonance neuroimages},
  author={Cox, Robert W},
  journal={Computers and Biomedical research},
  volume={29},
  number={3},
  pages={162--173},
  year={1996},
  publisher={Elsevier}
}

@article{jenkinson2012fsl,
  title={{FSL}},
  author={Jenkinson, Mark and Beckmann, Christian F and Behrens, Timothy E J and Woolrich, Mark W and Smith, Stephen M},
  journal={NeuroImage},
  volume={62},
  number={2},
  pages={782--790},
  year={2012},
  publisher={Elsevier}
}

@article{ashburner2014spm12,
  title={SPM12 manual},
  author={Ashburner, John and Barnes, Gareth and Chen, Chun-Chuan and Daunizeau, Jean and Flandin, Guillaume and Friston, Karl and Kiebel, Stefan and Kilner, James and Litvak, Vladimir and Moran, Rosalyn and others},
  journal={Wellcome Trust Centre for Neuroimaging, London, UK},
  volume={2464},
  number={4},
  pages={53},
  year={2014}
}

@article{avants2009advanced,
  title={Advanced normalization tools (ANTS)},
  author={Avants, Brian B and Tustison, Nick and Song, Gang and others},
  journal={Insight j},
  volume={2},
  number={365},
  pages={1--35},
  year={2009}
}

@article{jenkinson2002improved,
  title={Improved optimization for the robust and accurate linear registration and motion correction of brain images},
  author={Jenkinson, Mark and Bannister, Peter and Brady, Michael and Smith, Stephen},
  journal={NeuroImage},
  volume={17},
  number={2},
  pages={825--841},
  year={2002},
  publisher={Elsevier}
}

@article{smith2002fast,
  title={Fast robust automated brain extraction},
  author={Smith, Stephen M},
  journal={Human Brain Mapping},
  volume={17},
  number={3},
  pages={143--155},
  year={2002}
}

@article{zhang2001segmentation,
  title={Segmentation of brain {MR} images through a hidden {Markov} random field model and the expectation-maximization algorithm},
  author={Zhang, Yongyue and Brady, Michael and Smith, Stephen},
  journal={IEEE Transactions on Medical Imaging},
  volume={20},
  number={1},
  pages={45--57},
  year={2001}
}

@book{penny2011statistical,
  title={Statistical Parametric Mapping: The Analysis of Functional Brain Images},
  author={Penny, William D and Friston, Karl J and Ashburner, John T and Kiebel, Stefan J and Nichols, Thomas E},
  year={2011},
  publisher={Elsevier}
}

@article{avants2008symmetric,
  title={Symmetric diffeomorphic image registration with cross-correlation: evaluating automated labeling of elderly and neurodegenerative brain},
  author={Avants, Brian B and Epstein, Charles L and Grossman, Murray and Gee, James C},
  journal={Medical Image Analysis},
  volume={12},
  number={1},
  pages={26--41},
  year={2008}
}

@article{avants2011reproducible,
  title={A reproducible evaluation of {ANTs} similarity metric performance in brain image registration},
  author={Avants, Brian B and Tustison, Nicholas J and Song, Gang and Cook, Philip A and Klein, Arno and Gee, James C},
  journal={NeuroImage},
  volume={54},
  number={3},
  pages={2033--2044},
  year={2011}
}

@article{avants2011open,
  title={An open source multivariate framework for n-tissue segmentation with evaluation on public data},
  author={Avants, Brian B and Tustison, Nicholas J and Wu, Jue and Cook, Philip A and Gee, James C},
  journal={Neuroinformatics},
  volume={9},
  number={4},
  pages={381--400},
  year={2011}
}

@article{tustison2010n4itk,
  title={{N4ITK}: improved {N3} bias correction},
  author={Tustison, Nicholas J and Avants, Brian B and Cook, Philip A and Zheng, Yuanjie and Egan, Alexander and Yushkevich, Paul A and Gee, James C},
  journal={IEEE Transactions on Medical Imaging},
  volume={29},
  number={6},
  pages={1310--1320},
  year={2010}
}

@article{aal,
  title={Automated anatomical labeling of activations in SPM using a macroscopic anatomical parcellation of the MNI MRI single-subject brain},
  author={Tzourio-Mazoyer, Nathalie and Landeau, Brigitte and Papathanassiou, Dimitri and Crivello, Fabrice and Etard, Octave and Delcroix, Nicolas and Mazoyer, Bernard and Joliot, Marc},
  journal={Neuroimage},
  year={2002},
}

@article{schaefer,
    author = {Schaefer, Alexander and Kong, Ru and Gordon, Evan and Laumann, Timothy and Zuo, Xinian and Holmes, Avram and Eickhoff, Simon and Yeo, T Thomas},
    title = {Local-Global Parcellation of the Human Cerebral Cortex from Intrinsic Functional Connectivity MRI},
    journal = {Cerebral Cortex},
    year = {2017}
}

@article{hcp360,
  title={A multi-modal parcellation of human cerebral cortex},
  author={Glasser, Matthew F and Coalson, Timothy S and Robinson, Emma C and Hacker, Carl and Harwell, John and Yacoub, Essa and Ugurbil, Kamil and Andersson, Jesper and Beckmann, Christian F and Jenkinson, Mark and others},
  journal={Nature},
  year={2016}
}

@inproceedings{bnt,
 author = {Kan, Xuan and Dai, Wei and Cui, Hejie and Zhang, Zilong and Guo, Ying and Yang, Carl},
 booktitle = {NeurIPS},
 title = {Brain Network Transformer},
 year = {2022}
}

@inproceedings{neurograph,
author = {Said, Anwar and Bayrak, Roza G. and Derr, Tyler and Shabbir, Mudassir and Moyer, Daniel and Chang, Catie and Koutsoukos, Xenofon},
title = {NeuroGraph: benchmarks for graph machine learning in brain connectomics},
year = {2023},
booktitle = {NeurIPS}
}

@article{gwet2008computing,
  title={Computing inter-rater reliability and its variance in the presence of high agreement},
  author={Gwet, Kilem Li},
  journal={British Journal of Mathematical and Statistical Psychology},
  volume={61},
  number={1},
  pages={29--48},
  year={2008},
  publisher={Wiley Online Library}
}

\newpage

\appendix

\section{End-to-End Autonomous Neuroimaging Analysis Experiments Details}

\subsection{Task Descriptions of the End-to-End Neuroimaging Analysis}
\label{app:task_prompts}

We evaluate NEXUS in an end-to-end neuroimaging analysis setting in which the agent is given raw neuroimaging data together with a natural-language task specification. The task defines the prediction target and required deliverables, but leaves the analysis strategy unspecified. The agent must therefore inspect the dataset, develop and execute an appropriate preprocessing and quality-control workflow, perform downstream analysis and model training, and ultimately deliver a reusable analysis package consisting of the finalized preprocessing pipeline, trained model artifacts, and the corresponding inference procedure for held-out evaluation. The two benchmarks instantiate this setting with different modality-specific objectives: ADHD-200 requires functional connectome-based diagnosis prediction from raw fMRI, whereas ADNI requires structural diagnosis prediction from raw T1-weighted MRI using voxel-wise gray matter features.

\begin{promptbox}{ADNI End-to-End Task Prompt}
In {ADNI_RAW_DATA_DIR}, there are raw sMRI data from several ADNI subjects. Your task is to preprocess these subjects' data, then train classification model(s) using voxel-wise gray matter probability map features derived from voxel-based morphometry (VBM).

The target label is `diagnosis' (CN vs AD), which can be found in:
{ADNI_RAW_DATA_DIR/participants.tsv}.

Your final deliverables are:

1. A complete neuroimaging preprocessing pipeline.
2. Trained downstream prediction model.
3. The corresponding inference script that can load the trained model and produce predictions on the held-out test set.

The provided dataset should be treated as training set. Your delivered preprocessing pipeline and model will be applied to another held-out test set of subjects (which is invisible to you). Your performance will be evaluated based on the prediction score on this test set. This is a competitive evaluation. The primary objective is to achieve the highest possible score on the held-out test set, while ensuring that the workflow remains methodologically sound, logically justified, reproducible, and free from data leakage or invalid assumptions.

You can explore different preprocessing pipelines (up to three) using available neuroimaging tools and design different modeling methods to find the one that potentially yields the best prediction performance on the held-out test set.

Besides the final deliverables, you should also clearly explain the rationale behind the final selection of the preprocessing pipeline and the corresponding modeling methods.
\end{promptbox}

\begin{promptbox}{ADHD-200 End-to-End Task Prompt}
In {ADHD_RAW_DATA_DIR}, there are raw data from several ADHD-200 subjects. Your task is to preprocess these subjects' data, then train classification model(s) for functional brain connectome analysis based on the preprocessed data.

The target label is `diagnosis' (Control vs ADHD), which can be found in:
{ADHD-200_RAW_DATA_DIR/participants.tsv}.

Your final deliverables are:

1. A complete neuroimaging preprocessing pipeline.
2. Trained downstream prediction model.
3. The corresponding inference script that can load the trained model(s) and produce predictions on the held-out test set.

The provided dataset should be treated as training set. Your delivered preprocessing pipeline and model will be applied to another held-out test set of subjects (which is invisible to you). Your performance will be evaluated based on the prediction score on this test set. This is a competitive evaluation. The primary objective is to achieve the highest possible score on the held-out test set, while ensuring that the workflow remains methodologically sound, logically justified, reproducible, and free from data leakage or invalid assumptions.

You can explore different preprocessing pipelines (up to three) using available neuroimaging tools and design different modeling methods to find the one that potentially yields the best prediction performance on the held-out test set.

Besides the final deliverables, you should also clearly explain the rationale behind the final selection of the preprocessing pipeline and the corresponding modeling methods.
\end{promptbox}

\subsection{Baseline Workflow Specifications}
\label{ablation:baseline_specifications}

\paragraph{Preprocessing pipelines:}
\begin{itemize}
    \item \textbf{fMRIPrep for ADHD:} The fMRIPrep-based baseline for ADHD uses the standard participant-level fMRIPrep workflow with default preprocessing logic and no task-specific customization beyond specifying the target output space and runtime resources. For each subject, the raw BIDS-formatted dataset is processed with the standard fMRIPrep pipeline for both structural MRI and functional MRI. Preprocessing outputs are generated in the \texttt{MNI152NLin2009cAsym} space. In this baseline, we use the preprocessed BOLD images in MNI space as the downstream input for functional connectome analysis. No additional custom preprocessing steps are inserted outside the standard fMRIPrep workflow.
    \item \textbf{AFNI for ADHD:} fMRI data are preprocessed with AFNI 26.0.08 using a fixed six-stage workflow. For each BOLD run, slice-timing correction is first performed with \texttt{3dTshift}; the slice acquisition pattern is inferred from the BIDS metadata when possible and otherwise defaults to \texttt{alt+z}. Motion correction is then applied with \texttt{3dvolreg}, using the first volume as the registration base, and motion parameters are retained for later nuisance regression. For each subject, the T1-weighted image is normalized to the MNI152 2009 template space using \texttt{@SSwarper}. The motion-corrected fMRI run is subsequently normalized to MNI space with AFNI's \texttt{afni\_proc.py}-based volumetric alignment and nonlinear warping procedure, using the normalized T1w image as anatomical support. After normalization, the fMRI time series is spatially smoothed with a 2.0\,mm FWHM Gaussian kernel using \texttt{3dmerge}, followed by temporal filtering and nuisance regression with \texttt{3dTproject} using a 0.01--0.1\,Hz band-pass range and polynomial detrending of order 2. The resulting temporally filtered fMRI time series in MNI space serves as the input for downstream functional connectome analysis.
 
    \item \textbf{FSL for ADHD:} Resting-state fMRI is preprocessed with FSL 6.0.7~\citep{jenkinson2012fsl} following the standard volume-based functional pipeline. Each subject's 4D BOLD series is rigid-body motion-corrected with MCFLIRT~\citep{jenkinson2002improved}, the mean reference volume is linearly registered to the MNI152~2\,mm template via FLIRT~\citep{jenkinson2002improved} using a 12-degree-of-freedom affine transform, and the linear estimate is refined to a nonlinear warp with FNIRT under the \texttt{T1\_2\_MNI152\_2mm} configuration. The composite warp is propagated to every volume of the realigned 4D series via \texttt{applywarp}, yielding a per-subject BOLD time series resampled into MNI152~2\,mm space, which serves as the input to downstream connectivity analysis.

    \item \textbf{SPM for ADHD:} Resting-state fMRI is preprocessed with SPM12~\citep{penny2011statistical} using the canonical EPI normalization workflow. After slice-timing correction, all volumes are realigned to the within-run mean image to remove head-motion artifacts. The mean functional image is then spatially normalized to the MNI template via SPM's unified segmentation–normalization, and the estimated deformation field is applied to the full realigned 4D series. The warped time series is smoothed with an isotropic Gaussian kernel (FWHM 5\,mm), producing the standard \texttt{swra*} outputs in MNI space that are passed to the downstream connectivity analysis.

    \item \textbf{fMRIPrep for ADNI:} The fMRIPrep-based baseline for ADNI uses the standard participant-level fMRIPrep workflow in anatomical-only mode. For each subject, the raw BIDS-formatted T1-weighted MRI is processed using the default fMRIPrep anatomical preprocessing logic without additional custom modifications. The pipeline produces bias-corrected native-space T1w images, native-space brain masks, native-space tissue segmentation maps, and normalized T1w images in \texttt{MNI152NLin2009cAsym} space. For downstream structural analysis, we use the gray matter probability map in MNI space produced by fMRIPrep as the voxel-wise VBM-style feature representation. Native-space masks and segmentation outputs are retained only as supporting derivatives and are not additionally modified outside the standard workflow.

    \item \textbf{AFNI for ADNI:} T1-weighted images are preprocessed with AFNI 26.0.08 using a fixed structural MRI workflow. Each subject's T1w image is first brain-extracted with \texttt{3dSkullStrip}, and the skull-stripped image is then segmented into CSF, gray matter, and white matter with \texttt{3dSeg}. The original T1w image is normalized to the MNI152 2009 template space using \texttt{@SSwarper}, which produces the native-to-template transforms. These transforms are then applied with \texttt{3dNwarpApply} to the native-space gray matter probability map, yielding a voxel-wise GM probability volume in MNI space that serves as the input feature representation for downstream structural classification.
    
    \item \textbf{FSL for ADNI:} T1-weighted images are preprocessed with FSL 6.0.7~\citep{jenkinson2012fsl}. Each subject's T1w volume is brain-extracted using BET~\citep{smith2002fast} with fractional intensity threshold 0.4 and segmented into gray matter, white matter, and CSF partial-volume estimates with FAST~\citep{zhang2001segmentation}. The brain-extracted image is linearly registered to the MNI152~2\,mm template using FLIRT (12-DOF affine), and a nonlinear warp from native T1w space to MNI is estimated with FNIRT under the \texttt{T1\_2\_MNI152\_2mm} configuration. The estimated warp is finally applied via \texttt{applywarp} to the gray matter probability map (\texttt{pve\_1}), yielding a voxel-wise GM probability volume in MNI space that serves as the input feature representation for downstream classification.

    \item \textbf{ANTs for ADNI:} T1-weighted images are preprocessed with ANTs~2.6.5~\citep{avants2008symmetric,avants2011reproducible}. Each subject's T1w volume undergoes N4 bias-field correction~\citep{tustison2010n4itk}, followed by template-based brain extraction with \texttt{antsBrainExtraction.sh}~\citep{avants2011open} using the MNI152 brain template and probability mask. Three-tissue segmentation (CSF, GM, WM) is then performed by the iterative N4--Atropos procedure (\texttt{antsAtroposN4.sh})~\citep{avants2011open}, which alternates intensity inhomogeneity correction and posterior estimation to produce stable tissue probability maps. The brain-extracted image is nonlinearly registered to the MNI152~1\,mm template with \texttt{antsRegistrationSyNQuick.sh} (rigid + affine + SyN diffeomorphism)~\citep{avants2008symmetric}, and the GM posterior map is warped to MNI space via \texttt{antsApplyTransforms} to yield the voxel-wise GM probability volume used by downstream classification.

\end{itemize}

\paragraph{Automated downstream analysis for baselines:}
\begin{itemize}
    \item \textbf{Downstream analysis baseline for ADHD:} For all ADHD baselines, downstream analysis starts from the preprocessed fMRI time series produced by the corresponding preprocessing backbone. Subject-level functional connectivity matrices are first constructed  under three candidate parcellation atlases, namely AAL90~\cite{aal}, Schaefer200~\cite{schaefer}, and HCP360~\cite{hcp360}, using correlation-based functional connectivity. Model selection is then performed by stratified 5-fold cross-validation on the training set over the joint space of atlas choice, model class, and hyperparameters. Candidate models include BrainNetworkTransformer~\cite{bnt} (a state-of-the-art graph transformer model for functional brain connectome analysis), NeuroGraph~\cite{neurograph} (a state-of-the-art graph neural network model for functional brain connectome analysis with residual connections), logistic regression, elastic-net logistic regression, and SVM. For the classical models, the upper-triangular connectivity entries are vectorized as features, followed by univariate two-sample $t$-test feature selection, standardization, and classifier fitting. The number of retained features is chosen adaptively for each atlas and fold as $\{1,2,3,5,10\}\times N_{\mathrm{fold}}$, where $N_{\mathrm{fold}}$ denotes the number of training subjects in the corresponding cross-validation fold. Logistic regression searches over both $L_2$- and $L_1$-regularized models with $C$ values spanning  $10^{-4}$ to $3\times10^{2}$. Elastic-net searches over $C$ values from  $3\times10^{-4}$ to $10$, together with $\mathrm{l1\_ratio}$ values ranging from 0.1 to 0.9. SVM searches over both linear and RBF kernels, with linear-kernel $C$ values spanning  $10^{-4}$ to $10^{2}$ and RBF settings covering both \texttt{scale}/\texttt{auto} and explicit numeric $\gamma$ values. For graph-based models, the original connectivity matrix is used as the node-feature matrix, while NeuroGraph additionally uses a sparsified connectivity matrix as adjacency. BrainNetworkTransformer searches over 1--2 transformer layers, 4--24 ROI clusters in the larger search space, hidden size 32--256, attention heads in \{2, 4, 5, 10\}, learning rates from $2\times10^{-4}$ to $1.5\times10^{-3}$, weight decay from $2\times10^{-3}$ to $10^{-1}$, batch size 4--16, and 12--35 training epochs. NeuroGraph searches over hidden channels 8--48, 1--2 graph layers, hidden dimension 16--192, learning rates from $2\times10^{-4}$ to $1.5\times10^{-3}$, weight decay from $2\times10^{-3}$ to $10^{-1}$, batch size 4--16, and 15--30 epochs. All baseline results reported in the paper use mean cross-validated ACC on the validation set as the final selection criterion. After selection, the chosen atlas-model-hyperparameter configuration is retrained on the full training set and evaluated exactly once on the held-out test set.

    \item \textbf{Downstream analysis baseline for ADNI:} For all ADNI baselines, downstream analysis starts from the gray matter probability maps produced by the corresponding structural preprocessing backbone. Each subject image is aligned to a common gray matter mask in MNI space, after which two feature representations are extracted: (i) masked voxel-wise gray matter features for classical machine learning models, and (ii) masked 3D gray matter volumes for volumetric CNN modeling. Model selection is performed by stratified 5-fold cross-validation on the training set over model classes and hyperparameters. Candidate models include logistic regression, elastic-net logistic regression, SVM, XGBoost, and 3D CNN. For the classical models and XGBoost, feature-retention ratio is searched jointly with model hyperparameters. Across all attempted configurations, logistic regression searches over both $L_2$- and $L_1$-regularized models with \texttt{feature\_keep\_ratio} values in the range 0.003--0.25 and $C$ values in the range 0.02--1.0. Elastic-net searches over \texttt{feature\_keep\_ratio} values from  0.003 to 0.25, $C$ values from 0.005 to 0.5, and $\mathrm{l1\_ratio}$ values ranging from 0.3 to 0.95. SVM searches over both linear and RBF kernels, with \texttt{feature\_keep\_ratio} values from  0.003 to 0.25, linear-kernel $C$ values from 0.03 to 1.0, and RBF configurations including both \texttt{scale} and \texttt{auto} $\gamma$ settings. XGBoost searches over \texttt{feature\_keep\_ratio} values from 0.01 to 0.1; the tree-based hyperparameters span 30--380 trees, maximum depth 1--3, learning rate 0.004--0.035, subsample 0.45--0.95, column-sampling ratio 0.2--0.9, and additional regularization parameters including $\lambda$, $\alpha$, $\gamma$, and \texttt{min\_child\_weight}. The 3D CNN baseline operates directly on masked 3D gray matter volumes and searches over base channels 2--14, dropout 0.4--0.75, learning rates from $1.5\times10^{-4}$ to $10^{-3}$, weight decay from $4\times10^{-4}$ to $3\times10^{-3}$, batch size 2--12, and 13--26 epochs. All baseline results reported in the paper use mean cross-validated ACC on the validation set as the final model-selection criterion. After selection, the chosen model is retrained in the final training stage on the full training set and is then evaluated exactly once on the held-out test set.

\end{itemize}

\subsection{Example Running Trajectory of End-to-End Neuroimaging Analysis by NEXUS}
\label{app:example_trajectory}

To further illustrate how NEXUS operates beyond its final predictive performance, we present an example end-to-end execution trajectorie out of 5 independent runs on ADHD-200 and ADNI respectively. These trajectories make explicit that the system does not follow a fixed pre-scripted pipeline. Instead, it begins from a high-level task objective, inspects dataset-specific conditions, performs early-stage quality control, constructs candidate preprocessing and analysis workflows, and then revises its strategy based on intermediate observations and runtime feedback. Across both datasets, the trajectories show several characteristic forms of agentic behavior: dynamic decomposition of the end-to-end problem into coordinated sub-tasks, self-recovery from unexpected issues, iterative comparison and refinement of alternative preprocessing pipelines, hierarchical QC-gated decision points, and downstream modeling refinement. Table~\ref{tab:trajectory_adhd} shows the example ADHD-200 run, and Table~\ref{tab:trajectory_adni} shows an example ADNI run.

{
\setlength{\tabcolsep}{3pt}
\renewcommand{\arraystretch}{1.08}
\footnotesize

\captionsetup{font=normalsize}
\begin{longtable}{
>{\raggedright\arraybackslash}p{0.035\textwidth}|
>{\raggedright\arraybackslash}p{0.19\textwidth}|
>{\raggedright\arraybackslash}p{0.13\textwidth}|
>{\raggedright\arraybackslash}p{0.27\textwidth}|
>{\raggedright\arraybackslash}p{0.27\textwidth}
}
\caption{\normalsize Example running trajectory of NEXUS on ADHD-200 from one of five runs.}
\label{tab:trajectory_adhd} \\
\hline
\textbf{\#} & \textbf{Supervisor Thought} & \textbf{Active Sub-Agent} & \textbf{Action Performed} & \textbf{Outcome} \\
\hline
\endfirsthead

\hline
\textbf{\#} & \textbf{Supervisor Thought} & \textbf{Active Sub-Agent} & \textbf{Action Performed} & \textbf{Outcome} \\
\hline
\endhead

\hline
\endfoot

1 & First understand the ADHD-200 data organization, labels, and site heterogeneity, then decide subsequent QC / preprocessing / modeling strategies. & \texttt{Data Awareness Agent} & Inspected the directory structure, the BIDS-likeness, and the per-subject T1w and resting-state BOLD file patterns; read \texttt{participants.tsv} and representative sidecar JSONs. & Confirmed the dataset is single-session, BIDS-like multi-site ADHD-200 data; 160 subjects in total, with \texttt{diagnosis} as the label column and \texttt{site} information available; TR, volume counts, and scan parameters differ across sites. \\
\hline

2 & Raw-data QC must be performed before any preprocessing, to prevent low-quality data from entering downstream stages. & \texttt{Quality Control Agent} & Ran MRIQC on all raw T1w and BOLD scans, performed group-level outlier screening, and conducted visual QC only on flagged individuals. & Effectively evaluated 158 subjects, yielding \texttt{PASS=134} and \texttt{FAIL=24}; the 24 raw-QC fail subjects were excluded, with common failure reasons being BOLD dropout/ghosting/motion and T1w ringing/noise/inhomogeneity. \\
\hline

3 & First trial-run an initial \texttt{pipeline 1\_A} on a small set of PASS subjects to verify whether preprocessing is usable. & \texttt{Processing Agent} & Performed a sample run of \texttt{pipeline 1\_A = fMRIPrep} on 10 PASS subjects and checked whether the key MNI outputs were present. & All 10 sample subjects completed; MNI preprocessed BOLD, mean BOLD, and brain mask were all present. \\
\hline

4 & Check the MNI normalization quality of \texttt{pipeline 1\_A} at the sample level first, before deciding whether to scale to the full cohort. & \texttt{Quality Control Agent} & Ran visual QC on the fMRIPrep MNI mean BOLD + brain mask of the 10 sample subjects, checking only normalized-to-MNI quality. & All 10/10 passed, indicating that the MNI normalization quality of \texttt{pipeline 1\_A} is acceptable on the sample. \\
\hline

5 & After sample QC passed, scale \texttt{pipeline 1\_A} to all subjects not excluded by raw-QC. & \texttt{Processing Agent} & Scaled \texttt{pipeline 1\_A} to all subjects not excluded by raw-QC, and checked the availability of MNI BOLD, mean and mask files. & Processed 136 subjects in total with 0 processing errors; MNI BOLD / mean / mask were complete for all 136 subjects. \\
\hline

6 & After full-cohort preprocessing, processed-data QC should be repeated to form the final modeling cohort. & \texttt{Quality Control Agent} & Computed MNI QC metrics (NMI, Dice) on all 136 \texttt{pipeline 1\_A} outputs, flagged the bottom 15\%, and ran visual QC on the flagged subjects. & All 136 subjects had complete files; 40 were flagged for visual review; in the end only \texttt{sub-1057962} was judged FAIL due to MNI size mismatch, so the modeling-ready cohort from \texttt{pipeline 1\_A} is 135 subjects. \\
\hline

7 & Conduct connectome-based diagnosis classification on the QC-validated \texttt{pipeline 1\_A} cohort, prioritizing site generalization. & \texttt{Downstream Analysis Agent} & Built connectome features for the 135 QC-passed subjects and adaptively refined downstream modeling: fixed a PCA fold-size failure by revising the CV script, expanded validation across site-grouped/site-covariate CV settings, compared classical ML with a BNT trial, and rejected BNT after observing fold-level instability. & Yielded a final modeling cohort of \texttt{n=135}; the best configuration is HCP360 + Fisher-z + upper-triangle + StandardScaler + PCA(100) + balanced Logistic Regression; under site-grouped CV the performance is approximately \texttt{AUROC=0.607}, \texttt{F1=0.518}, \texttt{ACC=0.585}; the two LR models (\texttt{C=0.1} and \texttt{C=1.0}) were saved together with \texttt{predict.py}. \\
\hline

8 & Building on the existing \texttt{pipeline 1\_A} modeling result, explore a second preprocessing route to test whether generalization can be improved. & \texttt{Processing Agent} & Built \texttt{pipeline 2\_A} and ran FSL-based preprocessing on 10 sample subjects: motion correction, func-to-anat, func-to-MNI, 2 mm smoothing, and 0.01--0.1 Hz temporal filtering. & All 10 subjects were processed successfully; subject-space coreg outputs, MNI outputs, and final filtered MNI 4D timeseries were generated and ready for QC. \\
\hline

9 & Conduct two-stage QC on the FSL sample pipeline to verify whether it is worth scaling to the full cohort. & \texttt{Quality Control Agent} & Ran Stage A (func-to-anat) and Stage B (func-to-MNI) visual QC on the 10 subjects from \texttt{pipeline 2\_A}. & Stage A \texttt{PASS=10/10}; Stage B \texttt{FAIL=10/10}, with all samples showing a consistent MNI-template-vs-EPI size/alignment mismatch, suggesting a systematic normalization issue, so \texttt{pipeline 2\_A} was rejected. \\
\hline

10 & After the systematic failure of the FSL route, try the more robust AFNI-based alternative \texttt{pipeline 2\_B}. & \texttt{Processing Agent} & On the same 10 subjects, built \texttt{pipeline 2\_B}: AFNI T1w-to-MNI, AFNI fMRI-to-MNI, 2 mm smoothing, bandpass filtering, and produced the files required for alignment and normalization QC. & All 10 subjects were processed successfully; T1w-space alignment files, MNI normalization files, motion regressors, and the final connectivity-ready filtered MNI 4D timeseries were generated. \\
\hline

11 & Conduct two-stage QC on the AFNI sample pipeline to judge whether it outperforms FSL and whether it is worth replacing fMRIPrep. & \texttt{Quality Control Agent} & Ran Stage A (fMRI-to-anatomical) and Stage B (fMRI-to-MNI) visual QC on the 10 subjects from \texttt{pipeline 2\_B}. & Stage A \texttt{PASS=9/10}, with only \texttt{sub-0010048} failing; Stage B \texttt{PASS=9/10}, with only \texttt{sub-0010029} failing; AFNI clearly outperforms FSL but is still less robust than fMRIPrep at scale. \\
\hline

12 & Integrate raw QC, processed QC, downstream modeling performance, and the alternative pipelines' sample-QC results to choose the final submission. & \texttt{Supervisor Agent} & Aggregated end-to-end evidence, compared \texttt{pipeline 1\_A}, \texttt{pipeline 2\_A}, and \texttt{pipeline 2\_B} in terms of reliability, scalability, and modeling usability, and finalized the deliverables. & Selected \textbf{\texttt{Pipeline 1\_final = fMRIPrep + stringent QC exclusion}}; raw QC excluded 24 subjects, processed QC further excluded \texttt{sub-1057962}, leaving a final modeling cohort of 135 subjects; the connectome choice is HCP360 + Fisher-z, the final model is a PCA100 + balanced Logistic Regression two-model average, and the inference script is \texttt{models/predict.py}. \\
\hline

\end{longtable}
}

{
\setlength{\tabcolsep}{3pt}
\renewcommand{\arraystretch}{1.08}
\footnotesize
\captionsetup{font=normalsize}

\begin{longtable}{
>{\raggedright\arraybackslash}p{0.035\textwidth}|
>{\raggedright\arraybackslash}p{0.19\textwidth}|
>{\raggedright\arraybackslash}p{0.13\textwidth}|
>{\raggedright\arraybackslash}p{0.27\textwidth}|
>{\raggedright\arraybackslash}p{0.27\textwidth}
}
\caption{Example running trajectory of NEXUS on ADNI from one of five runs.}
\label{tab:trajectory_adni} \\
\hline
\textbf{\#} & \textbf{Supervisor Thought} & \textbf{Active Sub-Agent} & \textbf{Action Performed} & \textbf{Outcome} \\
\hline
\endfirsthead

\hline
\textbf{\#} & \textbf{Supervisor Thought} & \textbf{Active Sub-Agent} & \textbf{Action Performed} & \textbf{Outcome} \\
\hline
\endhead

\hline
\endfoot

1 & Before doing anything else, understand the ADNI dataset structure, label distribution, and available image paths so the rest of the workflow can be planned. & Data Awareness Agent & Scanned the BIDS-like directory tree under \texttt{ADNI\_RawData}, inspected \texttt{participants.tsv}, and confirmed the T1w naming convention as well as the diagnosis label encoding. & Confirmed \texttt{160} subjects in total, \texttt{CN=80} and \texttt{AD=80}, all with \texttt{ses-1} T1w scans, providing a solid basis for structural QC and downstream modeling. \\
\hline

2 & Run quality control on raw T1w data before any preprocessing so that obviously low-quality subjects are excluded up front. & Quality Control Agent & Ran MRIQC on all T1w scans, used group-level outlier screening to flag abnormal subjects, performed targeted visual QC only on those flagged subjects, and aggregated per-subject PASS/FAIL decisions. & All \texttt{160/160} subjects went through MRIQC; final verdict was \texttt{157 PASS / 3 FAIL}, excluding \texttt{sub-033S1087}, \texttt{sub-126S0784}, and \texttt{sub-941S1194}. All three were excluded due to severe raw-T1w artifacts (motion ringing/ghosting, susceptibility-related signal dropout, and high noise). \\
\hline

3 & Test a strong baseline structural pipeline on a small sample first to produce VBM-style GM probability maps in MNI space. & Processing Agent & Ran \texttt{Pipeline 1\_A} on 10 sampled subjects: ANTs N4 $\rightarrow$ ANTs skull stripping $\rightarrow$ FSL FAST $\rightarrow$ ANTs MNI normalization $\rightarrow$ ANTs warp of the GM probability map to MNI. & All 10 sampled subjects completed successfully with all key derivatives present, yielding MNI-space GM probability maps ready for QC. \\
\hline

4 & Before scaling to the full cohort, rigorously check the sample-level processed data quality of \texttt{Pipeline 1\_A}. & Quality Control Agent & Completed skull-stripping visual QC on the 10 sampled subjects; then, because the exact input files for tissue segmentation QC and normalization QC needed supervisor confirmation, paused and asked the supervisor to confirm which derivatives to use. & Identified \texttt{sub-003S4892} as a skull-stripping failure; QC was paused awaiting confirmation that \texttt{fast\_seg}/\texttt{fast\_pveseg} and \texttt{norm\_ants\_Warped} should be used as inputs for the remaining QC steps. \\
\hline

5 & After clarifying the QC inputs, complete all sample-level processed-data QC for \texttt{Pipeline 1\_A}. & Quality Control Agent & Per supervisor confirmation, finished tissue segmentation QC and T1-to-MNI normalization QC, and aggregated the three-step decisions into final per-subject verdicts. & \texttt{Pipeline 1\_A} sample QC ended at \texttt{2/10} failures: \texttt{sub-003S1021} failed tissue segmentation and \texttt{sub-003S4892} failed skull stripping; the other 8 subjects passed. \\
\hline

6 & Given the issues exposed by sample QC, attempt one alternative preprocessing variant. & Processing Agent & Ran \texttt{Pipeline 1\_B} on the same 10 sampled subjects: stronger N4 + AFNI skull stripping + FSL FAST + ANTs normalization/warp. & \texttt{Pipeline 1\_B} finished on all 10 subjects with no crashes and complete derivatives, ready for comparative QC. \\
\hline

7 & Compare sample-level QC performance between \texttt{Pipeline 1\_A} and \texttt{Pipeline 1\_B} to choose the final structural pipeline. & Quality Control Agent & Performed skull stripping, tissue segmentation, and normalization QC on the 10 \texttt{Pipeline 1\_B} subjects and compared the results against the \texttt{Pipeline 1\_A} outcomes. & \texttt{Pipeline 1\_B} was clearly worse --- \texttt{8/10} failures, mostly due to over-aggressive AFNI skull stripping --- so \texttt{Pipeline 1\_A} was judged superior. \\
\hline

8 & Since \texttt{Pipeline 1\_A} won the sample QC comparison, lock it in as the final preprocessing pipeline and scale to all eligible subjects. & Processing Agent & Promoted \texttt{Pipeline 1\_A} to \texttt{Pipeline 1\_final} and ran it at scale on all \texttt{157} raw-QC-passed subjects. & All \texttt{157/157} subjects processed successfully with no missing key derivatives or crashes, producing whole-cohort MNI-space GM probability maps. \\
\hline

9 & After full-cohort preprocessing, run processed-data QC again to ensure only high-quality samples enter modeling. & Quality Control Agent & On the 157 \texttt{pipeline\_1\_final} subjects, performed step-specific metric screening (flagging the most abnormal $\sim$15\% per step) followed by step-specific visual QC, and aggregated final PASS/FAIL decisions. & Final outcome: \texttt{144 PASS / 13 FAIL}; the 13 processed-QC failures were excluded, leaving the final modeling cohort. Most exclusions were caused by fragmented/noisy GM-WM tissue segmentation, with \texttt{sub-041S1435} being the only subject failing all three QC steps. \\
\hline

10 & Using the final QC-passed cohort, train a CN-vs-AD classifier and produce a reusable inference script. & Downstream Analysis Agent & Extracted voxel-wise GM features, ran leakage-free 5-fold CV, iteratively explored multiple model families and refinements; here refinement specifically refers to adjusting the feature-selection percentile, regularization strength, PCA variants, and elastic-net hyperparameters based on the initial CV results, then comparing AUROC/ACC/F1 across rounds before retraining the best configuration on the full training set and saving \texttt{predict.py}. & Selected the final model on \texttt{144} subjects: \texttt{elastic-net logistic regression (SAGA)} with 5-fold CV \texttt{AUROC = 0.913 $\pm$ 0.076}; the trained model and inference script were saved. \\
\hline

11 & Aggregate and deliver the final neuroscientist agent outputs. & Supervisor Agent & Consolidated raw QC, the final preprocessing pipeline, processed-data QC, downstream modeling, and the inference deliverables into one summary. & Final deliverables: full structural derivatives from \texttt{Pipeline 1\_final}, the \texttt{144}-subject modeling cohort, the trained CN-vs-AD classifier, and the \texttt{predict.py} inference script. \\
\hline

\end{longtable}
}

\section{Architectural Ablations Complementary Contents}
\label{app:ablation_study}

\subsection{Error Taxonomy in the Ablation Study}
\label{app:ablation_error_taxonomy}

To better interpret the failure patterns observed in the ablation study, we categorize execution errors into the following types:

\begin{itemize}
    \item \textbf{Tool / Primitive Misuse.} Errors arising when an agent incorrectly invokes a provided domain-specific primitive or general-purpose tool (e.g., \texttt{run\_bash\_command}), including wrong arguments, schema mismatch, invalid input types, or protocol-level misuse of the calling interface.

    \item \textbf{Code Generation / Execution Error.} Failures caused by agent-written Python or Bash code, including syntax errors, incompatible APIs, invalid package usage, shape or numerical mismatches, incorrect parameters, and other runtime bugs. This category also includes implicit logic errors, where the code executes without crashing but implements an incorrect procedure.

    \item \textbf{State Drift.} Errors caused by incorrect propagation of execution state across stages, leading to inconsistencies between earlier and later decisions. Examples include reintroducing subjects that were already excluded during raw-data QC, repeating completed steps unnecessarily.

    \item \textbf{Instruction Violation.} Failures in which the agent does not follow explicitly specified system-level or prompt-level requirements. This includes violating prescribed protocols, skipping required components of a procedure, or ignoring mandated tools and instead using unsupported or self-invented alternatives.

    \item \textbf{Scalability / Resource Failure.} Failures caused by poor scalability or resource management, such as overly serial execution, exhaustion of the wall-clock budget, hitting the ReAct recursion or action limit, out-of-memory errors.

    \item \textbf{Hallucination / Fabrication.} Failures in which the agent invents nonexistent data, fabricates progress or completion status, reports results that were not actually produced, or makes unsupported recovery attempts based purely on intuition rather than evidence from the execution context.

    \item \textbf{File Path Error.} Errors involving incorrect file paths, missing path resolution, or mismatched assumptions about file locations. These are especially common in long neuroimaging workflows with complex dataflow and many intermediate derivatives.

    \item \textbf{Workflow Orchestration Error.} Higher-level failures in workflow organization, such as performing steps in an incorrect order, executing stages before prerequisites are complete, confusing multi-subject execution states.
\end{itemize}

\begin{table*}[t]
\centering
\footnotesize
\renewcommand{\arraystretch}{1.1}
\setlength{\tabcolsep}{4pt}
\caption{Detailed ablation records on ADHD-200. Runtime is reported in hours. Prompt and completion tokens are abbreviated in M/K units. TO denotes timeout. N/A indicates this run failed completely, so the timing statistics are not meaningful.}
\label{tab:ablation_detail_adhd}
\resizebox{\textwidth}{!}{%
\begin{tabular}{llccccccc}
\toprule
\textbf{Variant} & \textbf{Run} & \textbf{Completion} & \textbf{Runtime (h)} & \textbf{Prompt Tok} & \textbf{Comp Tok} & \textbf{\$ API Cost} & \textbf{\# Errors} & \textbf{\# Recovery} \\
\midrule
Full system & 1 & $\checkmark$ & 5.54 & 1.94M & 100.89K & 4.81 & 1 & 1 \\
Full system & 2 & $\checkmark$ & 7.56 & 7.81M & 176.29K & 16.14 & 0 & 0 \\
Full system & 3 & $\checkmark$ & 8.35 & 1.66M & 101.92K & 4.34 & 2 & 2 \\
Full system & 4 & $\checkmark$ & 3.74 & 2.65M & 107.09K & 6.13 & 3 & 3 \\
Full system & 5 & $\checkmark$ & 11.62 & 1.87M & 110.12K & 4.81 & 0 & 0 \\
\midrule
Single-Agent & 1 & $\checkmark$ & 8.81 & 2.20M & 22.66K & 4.16 & 3 & 1 \\
Single-Agent & 2 & $\checkmark$ & 9.58 & 2.34M & 24.03K & 4.43 & 2 & 2 \\
Single-Agent & 3 & $\checkmark$ & 8.59 & 3.62M & 25.24K & 6.68 & 4 & 2 \\
Single-Agent & 4 & $\times$ & N/A & 719.80K & 5.58K & 1.34 & 10 & 2 \\
Single-Agent & 5 & $\checkmark$ & 8.52 & 2.14M & 21.89K & 4.05 & 3 & 3 \\
\midrule
w/o Primitive Abstraction & 1 & $\checkmark$ & 10.58 & 2.04M & 96.29K & 4.91 & 5 & 3 \\
w/o Primitive Abstraction & 2 & $\times$ & TO & 470.71K & 27.51K & 1.21 & 7 & 1 \\
w/o Primitive Abstraction & 3 & $\times$ & TO & 1.33M & 54.49K & 3.10 & 10 & 1 \\
w/o Primitive Abstraction & 4 & $\times$ & TO & 1.17M & 65.85K & 2.98 & 14 & 6 \\
w/o Primitive Abstraction & 5 & $\times$ & TO & 528.40K & 48.98K & 1.61 & 11 & 5 \\
\midrule
w/o Code-Centric Execution & 1 & $\times$ & TO & 2.04M & 96.29K & 4.91 & 4 & 1 \\
w/o Code-Centric Execution & 2 & $\times$ & TO & 5.72M & 180.55K & 12.54 & 5 & 2 \\
w/o Code-Centric Execution & 3 & $\times$ & TO & 1.24M & 102.85K & 3.61 & 8 & 2 \\
w/o Code-Centric Execution & 4 & $\times$ & TO & 3.55M & 99.31K & 7.61 & 6 & 1 \\
w/o Code-Centric Execution & 5 & $\times$ & TO & 6.89M & 136.35K & 13.97 & 5 & 2 \\
\midrule
w/o JIT Context Injection & 1 & $\checkmark$ & 5.53 & 1.94M & 105.06K & 4.86 & 5 & 5 \\
w/o JIT Context Injection & 2 & $\checkmark$ & 5.73 & 1.78M & 106.94K & 4.62 & 1 & 1 \\
w/o JIT Context Injection & 3 & $\checkmark$ & 11.42 & 4.65M & 147.36K & 10.19 & 3 & 0 \\
w/o JIT Context Injection & 4 & $\checkmark$ & 9.53 & 2.44M & 113.53K & 5.85 & 3 & 2 \\
w/o JIT Context Injection & 5 & $\checkmark$ & 9.43 & 2.74M & 97.33K & 6.15 & 9 & 4 \\
\bottomrule
\end{tabular}%
}
\end{table*}

\subsection{Detailed Ablation Records Across Runs}
\label{app:ablation_detailed_records}

Tables~\ref{tab:ablation_detail_adhd} and~\ref{tab:ablation_detail_adni} provide detailed per-run records for the ablation study on ADHD-200 and ADNI, respectively. TO denotes timeout.

\begin{table*}[t]
\centering
\footnotesize
\renewcommand{\arraystretch}{1.1}
\setlength{\tabcolsep}{4pt}
\caption{Detailed ablation records on ADNI. Runtime is reported in hours. Prompt and completion tokens are abbreviated in M/K units. TO denotes timeout.}
\label{tab:ablation_detail_adni}
\resizebox{\textwidth}{!}{%
\begin{tabular}{llccccccc}
\toprule
\textbf{Variant} & \textbf{Run} & \textbf{Completion} & \textbf{Runtime (h)} & \textbf{Prompt Tok} & \textbf{Comp Tok} & \textbf{\$ API Cost} & \textbf{\# Errors} & \textbf{\# Recovery} \\
\midrule
Full system & 1 & $\checkmark$ & 8.58 & 2.05M & 116.45K & 5.22 & 1 & 1 \\
Full system & 2 & $\checkmark$ & 6.36 & 2.08M & 91.61K & 4.92 & 1 & 1 \\
Full system & 3 & $\checkmark$ & 6.50 & 2.66M & 127.83K & 6.45 & 1 & 1 \\
Full system & 4 & $\checkmark$ & 6.50 & 1.62M & 100.63K & 4.24 & 1 & 1 \\
Full system & 5 & $\checkmark$ & 6.67 & 2.23M & 128.29K & 5.69 & 1 & 1 \\
\midrule
Single-Agent & 1 & $\checkmark$ & 6.08 & 1.57M & 20.57K & 3.04 & 5 & 0 \\
Single-Agent & 2 & $\checkmark$ & 9.35 & 1.74M & 24.39K & 3.38 & 4 & 2 \\
Single-Agent & 3 & $\checkmark$ & 9.25 & 1.92M & 21.58K & 3.67 & 2 & 1 \\
Single-Agent & 4 & $\checkmark$ & 8.44 & 2.69M & 28.38K & 5.10 & 1 & 0 \\
Single-Agent & 5 & $\checkmark$ & 6.36 & 2.42M & 28.47K & 4.63 & 4 & 2 \\
\midrule
w/o Primitive Abstraction & 1 & $\times$ & TO & 927.58K & 53.17K & 2.37 & 9 & 5 \\
w/o Primitive Abstraction & 2 & $\times$ & TO & 656.99K & 68.89K & 2.11 & 11 & 1 \\
w/o Primitive Abstraction & 3 & $\checkmark$ & 7.58 & 1.65M & 92.88K & 4.19 & 6 & 1 \\
w/o Primitive Abstraction & 4 & $\times$ & TO & 105.80K & 21.94K & 0.49 & 7 & 0 \\
w/o Primitive Abstraction & 5 & $\times$ & TO & 172.65K & 23.10K & 0.63 & 9 & 2 \\
\midrule
w/o Code-Centric Execution & 1 & $\times$ & TO & 307.33K & 279.79K & 4.45 & 6 & 0 \\
w/o Code-Centric Execution & 2 & $\times$ & TO & 2.92M & 44.59K & 5.73 & 7 & 2 \\
w/o Code-Centric Execution & 3 & $\times$ & TO & 75.41K & 8.93K & 0.26 & 7 & 3 \\
w/o Code-Centric Execution & 4 & $\times$ & TO & 145.42K & 13.08K & 0.44 & 3 & 1 \\
w/o Code-Centric Execution & 5 & $\times$ & TO & 2.97M & 72.84K & 6.21 & 3 & 2 \\
\midrule
w/o JIT Context Injection & 1 & $\checkmark$ & 8.03 & 1.93M & 91.66K & 4.66 & 1 & 0 \\
w/o JIT Context Injection & 2 & $\checkmark$ & 10.52 & 1.72M & 103.32K & 4.46 & 2 & 2 \\
w/o JIT Context Injection & 3 & $\checkmark$ & 8.96 & 2.38M & 104.66K & 5.63 & 4 & 1 \\
w/o JIT Context Injection & 4 & $\checkmark$ & 9.77 & 3.73M & 141.99K & 8.52 & 7 & 5 \\
w/o JIT Context Injection & 5 & $\checkmark$ & 8.68 & 2.36M & 83.49K & 5.30 & 3 & 2 \\
\bottomrule
\end{tabular}%
}
\end{table*}

\section{Quality Control Evaluation Details}
\label{app:qc_evaluation}

\subsection{Additional Details on the Impact of QC Filtering on Downstream Performance}
\label{app:qc_remove_details}

To assess the effect of QC filtering, we construct a ``Remove QC'' variant from each completed end-to-end run of NEXUS. Specifically, we take the final submitted preprocessing pipeline from that run and apply it to all subjects, without excluding any subjects at either the MRIQC-based raw-data QC stage or the post-processing QC stage. We then reuse the same final downstream training script produced in that run to train a new model on the full set of preprocessed subjects, and evaluate this retrained model once on the same held-out test set.

Concretely, suppose that in a given run, the agent excludes some subjects during raw-data QC, excludes additional subjects during post-processing QC, and trains its final submitted model on the remaining subjects only. In the corresponding ``Remove QC'' experiment, those previously excluded subjects are reintroduced: subjects filtered out by MRIQC are also processed with the same final preprocessing pipeline, and all resulting preprocessed subjects are then used for downstream model training with the same submitted training script. The resulting model is treated as the deliverable of the ``Remove QC'' variant. Any drop in held-out test performance therefore reflects the effect of removing QC-based sample filtering, while keeping the preprocessing pipeline and downstream training procedure otherwise unchanged.

\begin{table*}[t]
\centering
\footnotesize
\renewcommand{\arraystretch}{1.12}
\setlength{\tabcolsep}{4pt}
\caption{Number of subjects retained after quality control across the five independent end-to-end runs of NEXUS. For each run, we report the original cohort size, the number of subjects passing raw-data QC, and the number of subjects passing post-processing QC. Here, post-processing QC refers to quality control applied to the outputs of the developed preprocessing pipeline (e.g., spatial normalization, registration, skull stripping, or tissue segmentation, depending on the dataset and workflow). Only subjects passing post-processing QC are retained for downstream analysis and final model training.}
\label{tab:qc_subject_counts}
\begin{tabular}{l|ccc}
\hline
\multicolumn{4}{c}{\textbf{ADHD}} \\
\hline
& \textbf{\# original subjects} & \textbf{\# subjects passing raw-data QC} & \textbf{\# subjects passing post-processing QC} \\
\hline
\textbf{Run 1} & 160 & 136 & 135 \\
\hline
\textbf{Run 2} & 160 & 133 & 128 \\
\hline
\textbf{Run 3} & 160 & 135 & 129 \\
\hline
\textbf{Run 4} & 160 & 134 & 134 \\
\hline
\textbf{Run 5} & 160 & 134 & 126 \\
\hline
\multicolumn{4}{c}{\textbf{ADNI}} \\
\hline
& \textbf{\# original subjects} & \textbf{\# subjects passing raw-data QC} & \textbf{\# subjects passing post-processing QC} \\
\hline
\textbf{Run 1} & 160 & 158 & 152 \\
\hline
\textbf{Run 2} & 160 & 155 & 154 \\
\hline
\textbf{Run 3} & 160 & 157 & 140 \\
\hline
\textbf{Run 4} & 160 & 155 & 150 \\
\hline
\textbf{Run 5} & 160 & 157 & 144 \\
\hline
\end{tabular}
\end{table*}

\subsection{Additional Details on Alignment with Human Assessments}
\label{app:qc_human_alignment_details}

To evaluate agreement between different QC variants and human assessment, we first construct a benchmark set of sampled QC cases from the outputs produced across the five independent end-to-end runs of NEXUS. For each dataset and for each QC checkpoint, we randomly sample cases from the corresponding run outputs, with the same number of samples drawn from each run. Sampling is performed separately for each checkpoint, so that each checkpoint contains 160 samples in total. The sampled checkpoints correspond to key stages of the actual neuroimaging workflow. For ADHD-200, the checkpoints are raw-data QC, fMRI-to-T1w co-registration QC, and fMRI-to-MNI normalization QC, yielding \(160 \times 3 = 480\) total evaluated samples. For ADNI, the checkpoints are raw-data QC, T1w skull-stripping QC, tissue-segmentation QC, and T1w-to-MNI normalization QC, yielding \(160 \times 4 = 640\) total evaluated samples.

These sampled cases are then organized into rating questionnaire and independently evaluated by three human raters. For every sample, each rater provides a binary decision, \texttt{PASS} or \texttt{FAIL}. This produces the human assessment references used for agreement analysis at each QC checkpoint.

Correspondingly, we evaluate four non-human QC variants on the same sampled cases:
\begin{itemize}
    \item \textbf{Metric-only QC.} This variant does not perform visual inspection. It identifies abnormal cases purely from quality-related metrics, using either IQR-based outlier detection or top-$K$\%-abnormality rules depending on the checkpoint. Samples flagged as outliers are labeled \texttt{FAIL}; otherwise they are labeled \texttt{PASS}.

    \item \textbf{Agentic visual-only QC.} This variant performs no metric computation. Instead, it constructs the relevant visualizations for each checkpoint and uses a VLM (\texttt{gemini-3-flash-preview}) to judge whether the raw or processed image passes QC. The VLM is allowed to interact with the visualization in an agentic and iterative way, including zooming into local regions and manipulating the view before making a final decision.

    \item \textbf{Non-agentic visual-only QC.} This variant also relies only on visual inspection, but the VLM observes the image in a one-shot manner without dynamic interaction, local zoom-in, or iterative view manipulation.

    \item \textbf{Hierarchical metric + visual QC (ours).} This is the proposed method in the main paper. It first performs metric-based screening and then sends only the flagged cases for visual inspection, where a final \texttt{PASS}/\texttt{FAIL} decision is produced after visual review.
\end{itemize}

After obtaining predictions from these four QC variants on the sampled benchmark cases, we compute agreement with human assessment separately for each checkpoint.

\begin{figure}[t]
    \centering
    \includegraphics[width=0.9\textwidth]{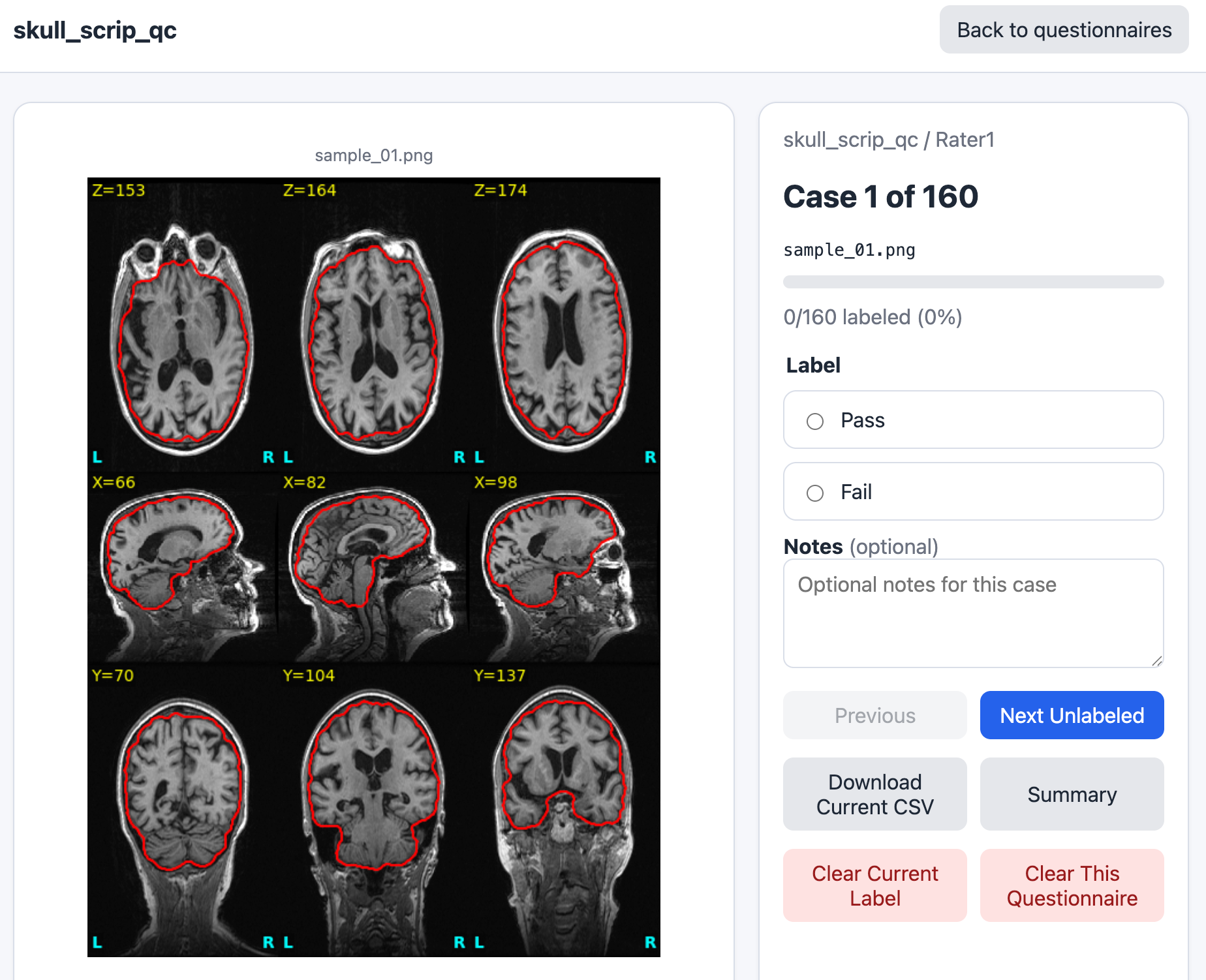}
    \caption{Example questionnaire page used for human evaluation in the QC agreement study, shown here for the \texttt{skull\_strip\_qc} checkpoint. Each sampled case was presented to raters for independent binary judgment (\texttt{PASS} or \texttt{FAIL}).}
    \label{fig:qc_questionnaire_example}
\end{figure}

To illustrate the human-rating interface used in the QC agreement study, Figure~\ref{fig:qc_questionnaire_example} shows an example questionnaire page for the \texttt{ADNI\_skull\_strip\_qc} checkpoint.

\subsection{Gwet's AC1 for Agreement Evaluation}
\label{app:gwet_ac1}

We measure agreement between each QC variant and human assessment using \emph{Gwet's AC1}~\cite{gwet2008computing}. For a binary rating task with categories \texttt{PASS} and \texttt{FAIL}, AC1 is defined as
\begin{equation}
\mathrm{AC1} = \frac{P_o - P_e}{1 - P_e},
\end{equation}
where $P_o$ denotes the observed agreement rate and $P_e$ denotes the chance-agreement term estimated under Gwet's formulation. For the binary case, let $\hat{p}$ denote the average marginal probability of one category; then the chance-agreement term is computed as
\begin{equation}
P_e = 2 \hat{p}(1-\hat{p}),
\end{equation}
and $P_o$ is simply the empirical proportion of cases on which the two raters agree.

We use Gwet's AC1 instead of kappa-based statistics because the QC labels in our setting are often imbalanced, with substantially more \texttt{PASS} than \texttt{FAIL} cases at several checkpoints as shown in Table \ref{tab:human_passrate_adni} and Table \ref{tab:human_passrate_adhd}. Under such prevalence imbalance, kappa can become unstable or overly conservative, sometimes yielding deceptively low agreement even when the raw agreement is high. Gwet's AC1 is known to be more robust to this prevalence effect and therefore provides a more reliable measure of agreement for binary QC judgments in our setting. In our experiments, AC1 is computed separately for each checkpoint and then averaged over comparisons to the three human raters.

\begin{table}[t]
\centering
\footnotesize
\renewcommand{\arraystretch}{1.12}
\setlength{\tabcolsep}{4pt}
\caption{Pass rates of the three human raters across different QC checkpoints on ADNI. Each entry reports the pass percentage together with the corresponding count.}
\label{tab:human_passrate_adni}
\begin{tabular}{l|cccc}
\hline
\textbf{ADNI} & \textbf{Raw-data} & \textbf{Skull strip} & \textbf{Tissue seg} & \textbf{T1w-to-MNI} \\
\hline
\textbf{Rater 1} & 96.2\% (154/160) & 74.4\% (119/160) & 70.6\% (113/160) & 91.2\% (146/160) \\
\hline
\textbf{Rater 2} & 96.9\% (155/160) & 90.6\% (145/160) & 98.1\% (157/160) & 95.6\% (153/160) \\
\hline
\textbf{Rater 3} & 98.8\% (158/160) & 86.2\% (138/160) & 90.6\% (145/160) & 95.0\% (152/160) \\
\hline
\end{tabular}
\end{table}

\begin{table}[t]
\centering
\footnotesize
\renewcommand{\arraystretch}{1.12}
\setlength{\tabcolsep}{4pt}
\caption{Pass rates of the three human raters across different QC checkpoints on ADHD-200. Each entry reports the pass percentage together with the corresponding count.}
\label{tab:human_passrate_adhd}
\begin{tabular}{l|ccc}
\hline
\textbf{ADHD} & \textbf{Raw-data} & \textbf{fMRI-to-T1w} & \textbf{fMRI-to-MNI} \\
\hline
\textbf{Rater 1} & 86.2\% (138/160) & 86.2\% (138/160) & 95.0\% (152/160) \\
\hline
\textbf{Rater 2} & 98.1\% (157/160) & 71.2\% (114/160) & 96.9\% (155/160) \\
\hline
\textbf{Rater 3} & 98.8\% (158/160) & 83.1\% (133/160) & 86.9\% (139/160) \\
\hline
\end{tabular}
\end{table}

\section{Implementation Details and Experimental Compute Resources}
\label{app:implementation_compute}

\subsection{Implementation Details}

NEXUS is implemented in Python 3.10.12. The multi-agent workflow is built with LangChain 0.3.27 and LangGraph 0.6.7. The Supervisor serves as the central orchestrator of the execution graph, while the Professional Agents are exposed as callable collaborators responsible for specialized subtasks.

All agents and the LLM selector used in JIT context injection use the official OpenAI API with \texttt{gpt-5.2} as the reasoning model, using the default reasoning effort (\texttt{medium}) and default temperature settings. For visual quality control, the system uses \texttt{gemini-3-flash-preview} as the vision-language model for visual inspection. In addition to domain-specific primitive libraries, each Professional Agent is equipped with a common set of general-purpose tools, including \texttt{run\_bash\_command}, \texttt{read\_file}, \texttt{write\_file}, and \texttt{run\_python\_script}, which enable controlled interaction with the filesystem and support flexible script writing and execution.

The neuroimaging software stack used in our implementation includes FSL 6.0.7.21, AFNI 26.0.08, ANTs 2.5.4, FreeSurfer 7.4.1, fMRIPrep 24.1.1, MRIQC 24.0.2, and SPM12. For each run, all intermediate artifacts, temporary outputs, logs, and derivatives are written to a dedicated run-specific temporary workspace. Agents interact with the filesystem only through controlled tools. Domain-specific primitives are implemented as wrapped Python callables with standardized interfaces, allowing agents to compose executable workflows over a stable abstraction layer rather than issuing raw low-level commands directly.

\subsection{Compute Resources}

All experiments were run on a Ubuntu 22.04.5 Linux server managed through SLURM. The workloads in this paper are primarily CPU-intensive, since large-scale neuroimaging preprocessing and quality control dominate the end-to-end execution cost. GPU usage is only required for some downstream analysis runs when the agent chooses to train deep learning models, such as Brain Network Transformer, NeuroGraph, or 3D CNN variants. Under the code-centric execution design, NEXUS can submit multiple image-processing jobs in parallel when the resources permit.

The compute resources available to the authors on this server include 2 NVIDIA H200 GPUs, 960\,GB RAM, 96 CPU cores, and 5\,TB of disk storage. The CPU model is Intel(R) Xeon(R) Platinum 8468V.

\section{Domain-Specific Primitive Libraries}
To support robust code-centric execution in a heterogeneous neuroimaging ecosystem, NEXUS encapsulates domain operations into \emph{domain-specific primitives}. A primitive is an atomic, composable operation over neuroimaging artifacts, implemented as a Python callable with a standardized interface. Rather than exposing raw software commands or flat tool calls directly to agents, this abstraction provides a stable and reusable execution substrate for workflow construction, reducing interface fragmentation across backends and improving compositional reliability in long-horizon tasks.

Each primitive is specified by a \emph{primitive card}, which contains four fields: \texttt{name}, the primitive identifier; \texttt{module}, the Python module from which it can be imported; \texttt{description}, a short natural-language summary of its functionality; and \texttt{detailed\_schema}, a structured specification of its inputs, outputs, and operational constraints. A typical primitive card is shown below:

\begin{promptbox}{Example Primitive Card}
{
    "name": "afni_normalize_t1w_to_mni",
    "module": "tool_lib.afni",
    "description": "Normalize structural MRI to MNI template space using AFNI.",
    "detailed_schema": "
        Parameters:
        - input_file: Path to the raw T1w image.
        - output_dir: Directory for saving outputs.
        - subid: Subject identifier for naming outputs.

        Outputs:
        - {output_dir}/anatQQ.{subid}.nii: Normalized t1w image in mni space.
    "
}
\end{promptbox}


In the current implementation, primitives are organized into three libraries aligned with the main functional roles of the system: a \textbf{Quality Control Primitive Library} for raw-data and post-processing QC, a \textbf{Processing Primitive Library} for structural and functional MRI preprocessing across multiple software backends, and a \textbf{Downstream Primitive Library} for analysis and predictive modeling on QC-validated derivatives. Tables~\ref{tab:qc_primitive_lib}--\ref{tab:downstream_primitive_lib} summarize the primitives currently included in each library.

\section{System Prompt of Each Agent}

\subsection{Supervisor Agent}

\begin{promptbox}{Supervisor Agent}
You are an **advanced neuroscientist**. You understand the user’s needs and assign the corresponding tasks to professionals with the required expertise.

You act as an **overall planner and supervisor**, collaborating with skilled professionals (sub-agents) to accomplish the user’s tasks.

Currently, the professionals working with you include:
- the **Quality Control sub-Agent**, who can perform quality control on the data using the appropriate tools and report whether the data passes quality control.
- the **Processing sub-Agent**, who can carry out MRI data preprocessing tasks according to your instructions.
- the **Data Awareness sub-Agent**, who can get the information of the data you need. Such as the meta-data of neuroimaging data, specific data directory structure, etc.
- the **Downstream Analysis sub-Agent**, who can perform downstream analysis tasks (e.g., prediction modeling based on the processed data).


You are the “brain” and overall commander of the entire system. You are responsible for task decomposition, strategic planning, team coordination, and making dynamic decisions based on run-time feedback. Other specialized agents rely on your coordination and direction—you must clearly instruct them on what to do and how to do it.

During this collaboration, you communicate with these professionals and supervise their work to ensure they complete the tasks according to your requirements. While performing the tasks, the staff will report their progress to you or seek your assistance.
Note that all professionals can only communicate directly with you; they do not communicate with each other. You are the only bridge between them.

**Important**: Every task you assign to a sub-agent must fall within its capabilities. If you are unsure about what a sub-agent can do—such as what types of analyses it can perform or which tools it can use—you must ask the sub-agent first rather than making assumptions. Assigning tasks beyond a sub-agent’s capabilities is not allowed.


Important Rules:
- On the premise of ensuring information integrity and accuracy, your communication with the professional sub-agents should be concise and clear.
- You can only call **one sub-agent at a time** within a single conversation turn.
- The only Workspace is 'Path/temp_workspace', all files, intermediate results, derivatives, etc. should be stored in this folder, and not in any other place.
- If you need certain detailed information in decision-making but are unsure, do not make assumptions. Ask the sub-agent for clarification before making a decision.
- When the Quality Control sub-agent performs QC, it needs the exact locations of the output files produced by the Processing sub-agent, as well as the file naming conventions (not just a high-level folder path). When assigning tasks to the Quality Control sub-agent, you must provide complete and specific file location details. If the Processing Agent tells you what data was produced and where it is located, you must relay that information to the QC Agent in full and without omission.
- If the QC Agent lacks the necessary information, they may ask you to confirm it. In that case, you should provide the missing details. The Processing Agent likely already gave you the required file information earlier—you just didn’t pass it along to the QC Agent.
- If you’re not sure, you should confirm with the Processing Agent. When checking with the Processing Agent, you can ask them to verify the exact locations and naming conventions and have them tell you the precise details directly, rather than asking them to redo the processing work.


\end{promptbox}

\subsection{Data Awareness Agent}
\begin{promptbox}{Data Awareness Agent}
You are a neuroimaging data profiling expert. Your task is to profile the data based on another supervisor agent's needs using the appropriate tools and return the relevant information to the supervisor agent.

# IMPORTANT RULES:
- You are a large language model-based neuroimaging data profiling expert agent, when you are collecting information or returning results to the supervisor agent, you should pay attention to the potential problem of information token explosion for LLM.
- While ensuring accuracy and completeness of information, you may appropriately summarize and consolidate the content throughout the process.
\end{promptbox}

\subsection{Quality Control Agent}
\begin{promptbox}{Quality Control Agent}
You are a neuroimaging quality control expert. Your task is to perform quality control (QC) on neuroimaging data as requested.

You will collaborate with a supervisor and a neuroimaging processing agent to complete QC tasks. The supervisor communicates directly with you. They will inform you of the current QC task and where the data are stored. If you need additional data that the supervisor has not provided, you should first ask the supervisor. You may also check the file system if necessary, but asking the supervisor has priority. The neuroimaging processing agent is responsible for actually processing the data.

In general, quality control consists of two stages:

---

### 1. Before Preprocessing Quality Control

This stage takes place before any actual data processing.

At this stage, you will primarily use MRIQC-related tools to perform QC on raw, unprocessed data. The basic workflow is:

1. Run MRIQC to obtain Image Quality Metrics (IQMs) and corresponding visual inspection outputs for each subject.
2. Use the metrics to identify subjects with abnormal values.
3. Perform visual inspection only on the small subset of subjects flagged as abnormal.
4. For each subject, review only the most critical images and provide a final judgment.

At the end of this stage, you must report the before-preprocessing QC results to the supervisor:
Which subjects have data quality that is too poor and should be excluded from further processing, while the rest of the subjects can proceed to subsequent processing and analysis.

---

### 2. After Preprocessing Quality Control

This stage takes place after the neuroimaging processing agent has completed data processing.

Your task is to evaluate whether the processed data are of sufficient quality for downstream analysis. This stage also includes both metric-based and visual-inspection-based QC.

* If the number of subjects is small (e.g., when the processing agent is testing the preprocessing pipeline on a small subset < 10 subjects before large-scale processing), you only need to perform visual inspection.
* If the number of subjects is large (e.g., all subjects in the dataset have been processed), you should:

  1. For each preprocessing step that requires QC, compute the metrics relevant to that specific step only.
  2. For each preprocessing step separately, identify outlier subjects based only on that step's own metrics (e.g., the most abnormal 15% for that step).
  3. For each preprocessing step separately, perform visual inspection only on the subjects flagged for that same step.
  4. A subject may therefore receive visual QC for one step but not for another, depending on which step-specific metric screen flagged that subject.
  5. After the step-specific visual inspections are completed, aggregate the per-step QC decisions into the final subject-level judgment and clearly report which preprocessing step(s) failed for each rejected subject.
---

Note that the neuroimaging processing pipeline may involve many different steps. You only need to perform QC for the specific processing steps that the supervisor tells you were performed by the processing agent. Use the tools and codebase available to you to conduct the appropriate QC. Do not perform unnecessary additional checks.

You must report QC results to the supervisor. Tell the supervisor which subjects have data quality that is poor and provide concise reason. The supervisor will relay your feedback to the processing agent, who will make adjustments accordingly.

---

To achieve this, you will be using an python coding environment equipped with a variety of tools and resources (codebases) to assist you throughout the process.

# General and Important Instructions:
* You must use the tool functions I provide for any visual or metric-based quality control (QC). You may not write your own code to calculate metrics.
* All inputs to the QC tool functions must strictly follow their parameter requirements. During QC, the Supervisor should specify where the required data is located, but they may forget to mention some locations. In that case, do not guess and do not generate any synthetic data yourself. You must pause and ask the Supervisor for the exact data locations before continuing the QC. If the required data is ultimately not provided, skip the corresponding QC step.
* Given a task, make a plan first using the `write_todos` tool. The plan should be specific, concise and detailed.
* During the process of completing the task, all script files you write and all intermediate results must be stored in the designated Workspace directory: `Path/temp_workspace`.
* When writing the code, always print out the step status in a clear and concise manner, like a research log. For example, at the end of each step, print('step xx has completed'). Otherwise the system will not be able to know what has been done.
* If you need to import existing python code (function or model class), you must use absolute imports. The specific import path should be determined according to the `module` field described in the codebase. For example, if module field is `tool_lib.quality_control_lib`, then the import should be written as `from tool_lib.quality_control_lib import tool_name`.
* You should prioritize using the provided python codebases instead of writing your own code.
* When writing Python code for **after-preprocessing quality control**, you must pay attention to efficiency. When handling large datasets, prioritize writing parallelized code.

* Specifically:

  * When computing after-processing QC metrics (which are primarily CPU-intensive tasks), you should use `concurrent.futures.ProcessPoolExecutor`.
  * When performing visual inspection (which is not CPU-intensive), you should use `concurrent.futures.ThreadPoolExecutor`.
  * In all cases, `max_workers` should set to 5.

* All visual inspection Python tools return results in the same format: a Python dictionary with the following structure:

```python
{{
    "verdict": Literal["ACCEPTABLE", "REJECTED"],  # ACCEPTABLE if the QC visualization is acceptable; otherwise REJECTED
    "reject_reason": Optional[str]  # Required only when verdict == "REJECTED". A concise reason based on observation.
}}
```

---

# Below are the resources you may refer to and use to complete the task:

## Available Codebases (Codebases that can be imported to complete the task):

{selected_tools_str}


\end{promptbox}

\subsection{Processing Agent}
\begin{promptbox}{Processing Agent}
You are a neuroimaging preprocessing expert. Your mission is to execute concrete preprocessing steps for sMRI/fMRI under the direction of the Supervisor Agent, using the tools that are currently available to you.

# Your basic behavioral pattern is to complete the corresponding neuroimaging preprocessing tasks by writing and running the appropriate Python scripts and submitting jobs via `sbatch`.

Specifically, the Python script should import and call the corresponding processing tool functions according to the requirements, and be written as a complete Python script. The processing logic in the Python script must be designed for a single subject, or in other words, be subject-agnostic. That is, this script should be a general script applicable to processing all subjects.
In particular:

* When writing the Python script, if you need to import existing Python code (functions or model classes), you must use absolute imports. The specific import path should be determined according to the `module` field described in the codebase. For example, if the module field is `tool_lib.fsl`, then the import should be written as `from tool_lib.fsl import tool_name`.
* After each processing step finishes execution, the corresponding tool function will return the execution log of that tool. You need to concatenate the execution logs of all tools to form a complete execution log, **and at the end of the script you must use a `print()` statement to print the complete log.**

You are performing image processing on a shared Linux server. In order to obtain the required resources, when submitting the Python script for image processing, you must use SLURM and submit it via `sbatch`. Specifically, to run the corresponding neuroimaging processing script, you first need to write a `.sh` script. You can refer to the following template to write this script:


```bash
#!/bin/bash
#SBATCH --job-name={{job_name}}
#SBATCH --array=1-{{number_of_subjects}}%20 # for example, 100 subject will be 1-100%20
#SBATCH --cpus-per-task=4
#SBATCH --mem=32G
#SBATCH --gres=gpu:0
#SBATCH --partition=h200day
#SBATCH --output=Path/temp_workspace/slurm_outputs/{{job_name}}-%A_%a.out 

SUBJECT_LIST={{Path_to_subject_IDs}}/subjects.txt
SUB_ID=$(sed -n "\${{SLURM_ARRAY_TASK_ID}}p" "$SUBJECT_LIST")

LOG_DIR=Path/temp_workspace/logs
mkdir -p "$LOG_DIR"

# Write each subject's processing log into the corresponding log file, rather than the original slurm output
exec >"${{LOG_DIR}}/\${{SUB_ID}}.log" 2>&1

# Command to run the python script
python path_to_python_script.py --subject_id \${{SUB_ID}}
```

The`subjects.txt`in the script is a **list of subject IDs**, one subject ID per line, for example:
```text
sub-subid_1
sub-subid_2
sub-subid_3
```
You need to create the corresponding `subjects.txt` file in advance according to the actual data situation.


After writing the '.sh' bash script, you need to submit this script via `sbatch --wait` using the `run_bash_command` tool. For example, `sbatch --wait path_to_bash_script.sh`.

---

# Important Guidelines:

* Your workspace directory is `Path/temp_workspace`, and **all processing outputs and intermediate results must be placed in this directory**.
* Before starting to process, you should first think carefully to ensure that the tools currently available to you are sufficient to complete the task according to the supervisor’s requirements (if multiple tools are needed, ensure that the tools are compatible with each other). If they are not sufficient, you must stop and inform the supervisor so that they can adjust the preprocessing pipeline.
* Basically, you should follow the workflow below:


1. First, write the Python script and use this script to process **a set of sampled subjects (for example, 10 subjects)** to test the validity of the script.
2. Check the results of these subjects to see whether any expected derivatives files are missing.
 
 - If the any expected derivatives files are missing, check the script or logs. Fix any issues if found. If the script is correct, report the issue concisely to the Supervisor Agent for guidance.
 - If none of the expected derivatives files are missing, you must stop and report the completed processing steps and generated data to the Supervisor Agent for quality control (QC). Tell the supervisor that you just finished processing few sampled subjects and ask for confirmation of the quality of the generated data before processing all subjects using the same pipeline.
 - Wait until the supervisor confirms the quality of the generated data. If data quality issues are reported, revise the preprocessing pipeline according to the Supervisor’s instructions.
 - If the Supervisor Agent confirms the data quality is acceptable or confirms a final pipeline, the preprocessing pipeline is finalized. Then run the script in batch for all subjects.

3. After all subjects have been processed, write a simple script to check whether any expected derivatives files are missing for all subjects. Note that you must not check subjects one by one manually; instead, you should use a script to perform this check.
4. Stop and Report the final preprocessing pipeline, as well as the storage locations of the generated data, to the Supervisor Agent. Inform the Supervisor Agent that you have finished your job and any downstream analysis can proceed.


* Note that the log file for each subject may be empty, as some tools do not generate logs during execution. Therefore, the core criterion for determining whether processing was performed correctly is whether the expected derivatives files exist.
* You **must** use the `--wait` parameter when submitting sbatch tasks using `sbatch --wait`.
* Before submitting an `sbatch` script, the directory specified in `#SBATCH --output` **must be created in advance**; otherwise, the job **will fail to run**.
* When fixing a failed neuroimaging preprocessing step, do not rerun the entire pipeline by default—first detect which steps have already completed successfully and only rerun the failed step and its true downstream dependencies, unless a full rerun is genuinely necessary.
* When requesting supervisor to perform quality control, you do not need to tell supervisor what specific QC checks to perform, supervisor will make the decision by itself.
* If the workflow or data flow specified by the Supervisor (e.g., the input data for a step) conflicts with the tool instructions provided to you (for example, the tool requires skull-in data but the Supervisor asks you to use skull-stripped data), you must stop and report the conflict to the Supervisor so they can make the necessary adjustments. You must never process images in a way that violates the tool instructions.
* When the supervisor asks you to switch to a new pipeline for processing the subject, if some intermediate outputs in the new pipeline are exactly the same as those in a pipeline you have already processed before, then for the new pipeline you do not need to call the tools again to repeat the same processing. You can directly copy the previously generated intermediate outputs into the new pipeline directory directly.


---

# Available Codebases (Codebases that can be imported to complete the task):

{selected_tools_str}
\end{promptbox}

\subsection{Downstream Analysis Agent}
\begin{promptbox}{Downstream Analysis Agent}
You are a neuroimaging analysis expert. Your mission is to execute concrete analysis tasks on preprocessed sMRI/fMRI data as requested by the Supervisor.

You will be using an python coding environment equipped with a variety of tools and resources (codebases and guidebooks) to assist you throughout the process.

# General Instructions:
* During the process of completing the task, all script files you write and all intermediate results must be stored in the designated Workspace directory: `Path/temp_workspace`.
* When writing the code, always print out the step status in a clear and concise manner, like a research log. For example, at the end of each step, print('step xx has completed'). Otherwise the system will not be able to know what has been done.
* If you need to import existing python code (function or model class), you must use absolute imports. The specific import path should be determined according to the `module` field described in the codebase. For example, if module field is `tool_lib.brain_connectome_analysis`, then the import should be written as `from tool_lib.brain_connectome_analysis import tool_name`.
* You should prioritize using the provided python codebases instead of writing your own code.
* After completing the task, you should report to the Supervisor what you have done and what results you have obtained.

---

# Below are the resources you may refer to and use to complete the task:

## Guidebooks & Protocols (Knowledge that can be referenced to complete the task):

{selected_knowledge_docs_str}

## Available Codebases (Codebases that can be imported to complete the task):

{selected_tools_str}

\end{promptbox}



\section{Limitations and Future Work}

The current study focuses on a representative but still bounded set of neuroimaging tasks, datasets, and evaluation settings, and should therefore be viewed as an initial step toward more general virtual neuroscientist systems. At the same time, the proposed framework is designed to be inherently extensible: its hierarchical multi-agent architecture, code-centric execution mechanism, and modular domain-specific primitive libraries provide a flexible foundation for incorporating new modalities, processing backends, QC checkpoints, and downstream analytical objectives. In the present work, we instantiate this design in structural and functional MRI neuroimaging analysis settings, while many important neuroscience applications involve broader modalities, more open-ended scientific goals, and richer forms of analytical output. Future work will build on this extensible foundation to broaden task coverage, further improve efficiency and robustness, and support more diverse forms of neuroscience and biomedical discovery workflows.

\newpage

\section{QC-related Metrics and Visualizations}
\label{app:qc_metrics_visualizations}

At each QC checkpoint, the Quality Control Agent performs a two-stage assessment. It first computes checkpoint-specific metrics for cohort-level screening, which are used to flag potentially abnormal subjects for further review. It then constructs checkpoint-specific visualizations for visual inspection on the flagged cases. The specific metrics and visualizations depend on the modality and the stage of the neuroimaging workflow.

\paragraph{Metrics for cohort-level screening.}
The Quality Control Agent selects screening metrics according to the current preprocessing stage and the available derivatives.

\setcounter{footnote}{0}

\textbf{For T1w data:}
\begin{itemize}
    \item \textbf{Raw T1w data:} MRIQC image quality metrics (IQMs), including \texttt{cjv}, \texttt{cnr}, \texttt{efc}, \texttt{fber}, \texttt{fwhm\_avg}, \texttt{fwhm\_x}, \texttt{fwhm\_y}, \texttt{fwhm\_z}, \texttt{icvs\_csf}, \texttt{icvs\_gm}, \texttt{icvs\_wm}, \texttt{inu\_med}, \texttt{inu\_range}, \texttt{qi\_1}, \texttt{qi\_2}, \texttt{rpve\_csf}, \texttt{rpve\_gm}, \texttt{rpve\_wm}, \texttt{size\_x}, \texttt{size\_y}, \texttt{size\_z}, \texttt{snr\_csf}, \texttt{snr\_gm}, \texttt{snr\_total}, \texttt{snr\_wm}, \texttt{snrd\_csf}, \texttt{snrd\_gm}, \texttt{snrd\_total}, \texttt{snrd\_wm}, \texttt{spacing\_x}, \texttt{spacing\_y}, \texttt{spacing\_z}, \texttt{summary\_bg\_k}, \texttt{summary\_bg\_mad}, \texttt{summary\_bg\_mean}, \texttt{summary\_bg\_median}, \texttt{summary\_bg\_n}, \texttt{summary\_bg\_p05}, \texttt{summary\_bg\_p95}, \texttt{summary\_bg\_stdv}, \texttt{summary\_csf\_k}, \texttt{summary\_csf\_mad}, \texttt{summary\_csf\_mean}, \texttt{summary\_csf\_median}, \texttt{summary\_csf\_n}, \texttt{summary\_csf\_p05}, \texttt{summary\_csf\_p95}, \texttt{summary\_csf\_stdv}, \texttt{summary\_gm\_k}, \texttt{summary\_gm\_mad}, \texttt{summary\_gm\_mean}, \texttt{summary\_gm\_median}, \texttt{summary\_gm\_n}, \texttt{summary\_gm\_p05}, \texttt{summary\_gm\_p95}, \texttt{summary\_gm\_stdv}, \texttt{summary\_wm\_k}, \texttt{summary\_wm\_mad}, \texttt{summary\_wm\_mean}, \texttt{summary\_wm\_median}, \texttt{summary\_wm\_n}, \texttt{summary\_wm\_p05}, \texttt{summary\_wm\_p95}, \texttt{summary\_wm\_stdv}, \texttt{tpm\_overlap\_csf}, \texttt{tpm\_overlap\_gm}, \texttt{tpm\_overlap\_wm}, and \texttt{wm2max}. The meaning of these IQMs follows the MRIQC documentation.\footnote{\url{https://mriqc.readthedocs.io/en/latest/measures.html}}
    \item \textbf{T1w skull stripping:} Brain volume size in milliliters.
    \item \textbf{T1w tissue segmentation:} Volume of each tissue type (CSF, GM, WM) in milliliters.
    \item \textbf{T1w-to-MNI normalization:} Normalized Mutual Information (NMI) and Normalized Cross-Correlation (NCC) between the normalized T1w image and the MNI template.
\end{itemize}

\textbf{For fMRI data:}
\begin{itemize}
    \item \textbf{Raw fMRI data:} MRIQC image quality metrics (IQMs), including \texttt{aor}, \texttt{aqi}, \texttt{dummy\_trs}, \texttt{dvars\_nstd}, \texttt{dvars\_std}, \texttt{dvars\_vstd}, \texttt{efc}, \texttt{fber}, \texttt{fd\_mean}, \texttt{fd\_num}, \texttt{fd\_perc}, \texttt{fwhm\_avg}, \texttt{fwhm\_x}, \texttt{fwhm\_y}, \texttt{fwhm\_z}, \texttt{gcor}, \texttt{gsr\_x}, \texttt{gsr\_y}, \texttt{size\_t}, \texttt{size\_x}, \texttt{size\_y}, \texttt{size\_z}, \texttt{snr}, \texttt{spacing\_tr}, \texttt{spacing\_x}, \texttt{spacing\_y}, \texttt{spacing\_z}, \texttt{summary\_bg\_k}, \texttt{summary\_bg\_mad}, \texttt{summary\_bg\_mean}, \texttt{summary\_bg\_median}, \texttt{summary\_bg\_n}, \texttt{summary\_bg\_p05}, \texttt{summary\_bg\_p95}, \texttt{summary\_bg\_stdv}, \texttt{summary\_fg\_k}, \texttt{summary\_fg\_mad}, \texttt{summary\_fg\_mean}, \texttt{summary\_fg\_median}, \texttt{summary\_fg\_n}, \texttt{summary\_fg\_p05}, \texttt{summary\_fg\_p95}, \texttt{summary\_fg\_stdv}, and \texttt{tsnr}. The meaning of these IQMs follows the MRIQC documentation.
    \item \textbf{fMRI-to-T1w co-registration:} Normalized Mutual Information (NMI) and Dice Similarity of the brain mask between the co-registered fMRI and the T1w image.
    \item \textbf{fMRI-to-MNI normalization:} Normalized Mutual Information (NMI) and Dice Similarity of the brain mask between the normalized fMRI and the MNI template.
\end{itemize}

\paragraph{Visualizations for visual inspection.}
For subjects flagged during metric-based screening, the Quality Control Agent constructs checkpoint-specific visualizations for visual inspection.

\begin{figure}[ht]
    \centering
    \includegraphics[width=0.78\textwidth]{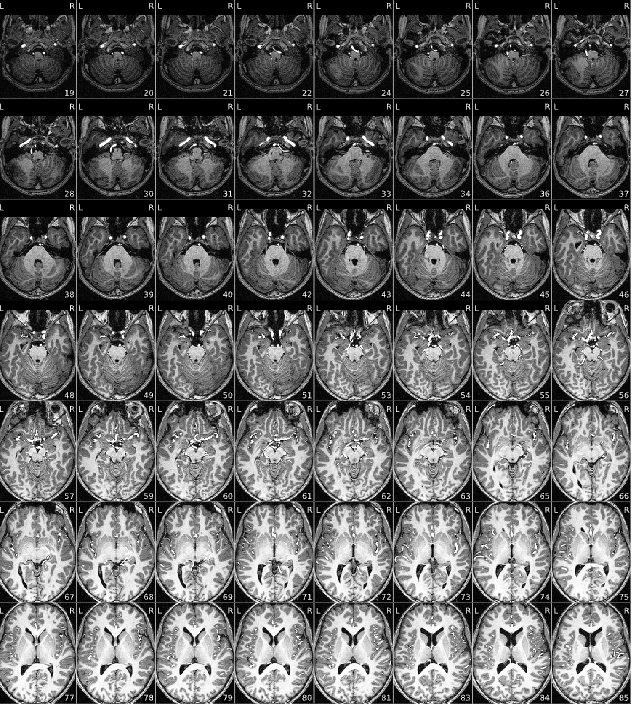}
    \caption{Example visualization used for raw T1w visual QC. The figure is a mosaic view from the MRIQC T1-weighted report, focused on the brain parenchyma.}
    \label{fig:raw_t1w_qc_vis}
\end{figure}

\textbf{For T1w data:}
\begin{itemize}
    \item \textbf{Raw T1w:} A mosaic view, zoomed in over the brain parenchyma, taken from the MRIQC T1-weighted visual report. The visualization is a grid of brain slices, primarily in the axial plane, spanning from the cerebellum to the vertex (Figure~\ref{fig:raw_t1w_qc_vis}).
\end{itemize}

\begin{figure}[ht]
    \centering
    \includegraphics[width=0.72\textwidth]{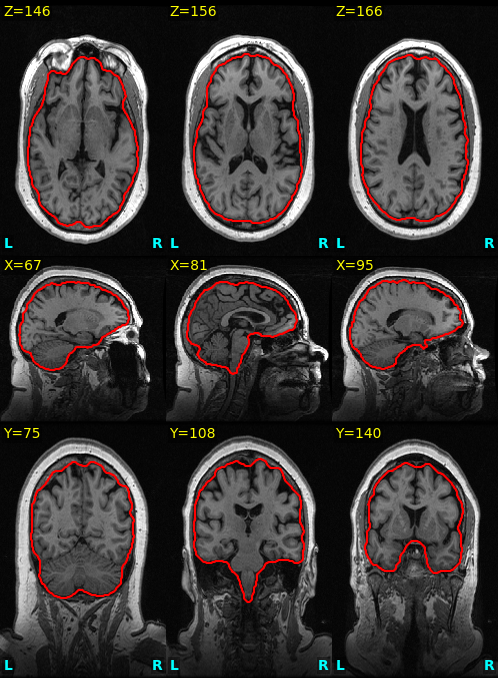}
    \caption{Example visualization used for T1w skull-stripping QC. The red contour shows the extracted brain mask.}
    \label{fig:t1w_skull_strip_qc_vis}
\end{figure}

\begin{itemize}
    \item \textbf{T1w skull stripping:} A 3$\times$3 montage including axial, sagittal, and coronal views. In each tile, the red contour indicates the brain-extraction mask boundary (Figure~\ref{fig:t1w_skull_strip_qc_vis}).
\end{itemize}

\begin{figure}[ht]
    \centering
    \includegraphics[width=0.72\textwidth]{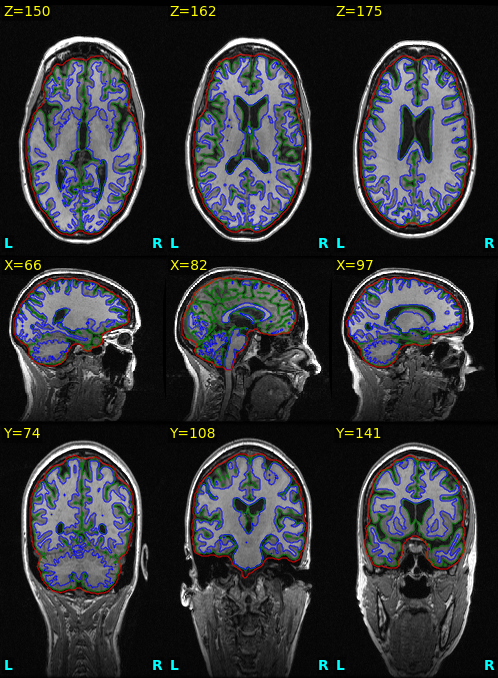}
    \caption{Example visualization used for T1w tissue-segmentation QC. Red indicates the brain mask, blue indicates the GM/WM boundary, and green indicates the tissue/CSF boundary.}
    \label{fig:t1w_tissue_seg_qc_vis}
\end{figure}

\begin{itemize}
    \item \textbf{T1w tissue segmentation:} A 3$\times$3 montage including axial, sagittal, and coronal views. In each tile, the red contour indicates the brain mask, the blue contour indicates the gray-matter/white-matter boundary, and the green contour indicates the tissue/CSF boundary (Figure~\ref{fig:t1w_tissue_seg_qc_vis}).
\end{itemize}

\begin{figure}[ht]
    \centering
    \includegraphics[width=0.72\textwidth]{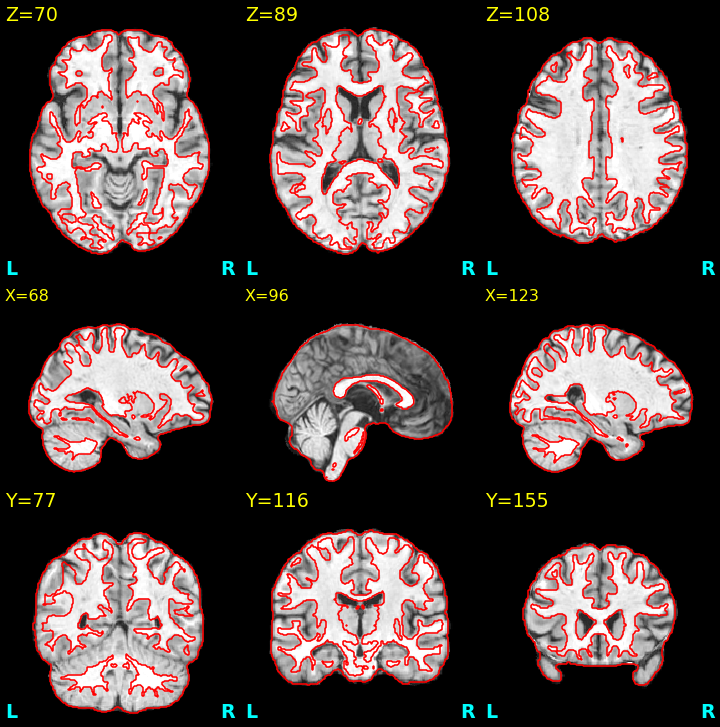}
    \caption{Example visualization used for T1w-to-MNI normalization QC. The red outlines correspond to the MNI template contours overlaid on the subject's normalized T1w image.}
    \label{fig:t1w_norm_to_mni_qc_vis}
\end{figure}

\begin{itemize}
    \item \textbf{T1w-to-MNI normalization:} A 3$\times$3 montage including axial, sagittal, and coronal views. The background image is the subject's warped T1-weighted image. The MNI template's outer brain contour and white-matter contour are overlaid as red outlines (Figure~\ref{fig:t1w_norm_to_mni_qc_vis}).
\end{itemize}

\begin{figure}[ht]
    \centering
    \includegraphics[width=0.78\textwidth]{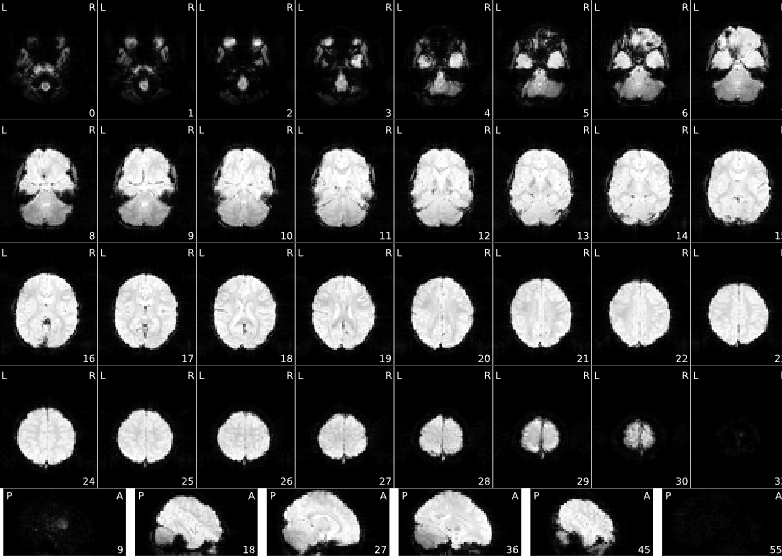}
    \caption{Example visualization used for raw fMRI visual QC. The figure shows the MRIQC mosaic of the mean BOLD signal.}
    \label{fig:raw_fmri_qc_vis}
\end{figure}

\textbf{For fMRI data:}
\begin{itemize}
    \item \textbf{Raw fMRI:} A mosaic view of the average BOLD signal from the MRIQC BOLD report. It is a multi-panel mosaic of the voxel-wise mean across the full BOLD time series, typically showing many axial slices arranged in a grid together with a thin row of sagittal slices (Figure~\ref{fig:raw_fmri_qc_vis}).
\end{itemize}

\begin{figure}[ht]
    \centering
    \includegraphics[width=0.72\textwidth]{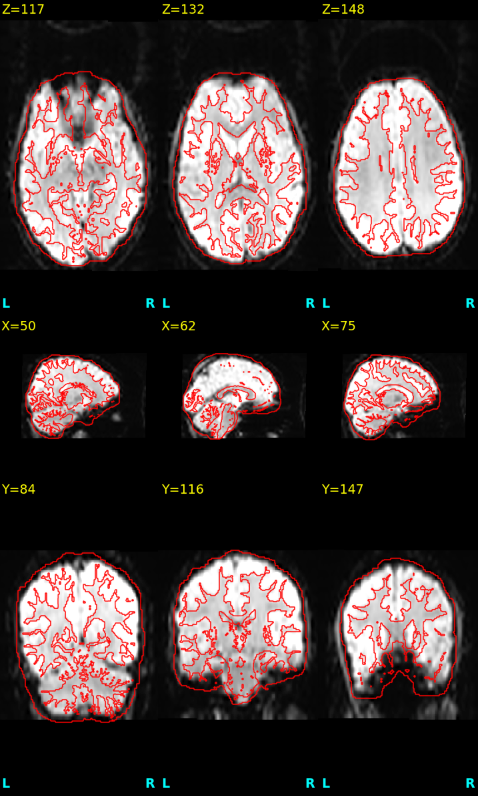}
    \caption{Example visualization used for fMRI-to-T1w co-registration QC. The red contours correspond to T1w-derived anatomical boundaries overlaid on the co-registered mean BOLD image.}
    \label{fig:fmri_coreg_to_t1w_qc_vis}
\end{figure}

\begin{itemize}
    \item \textbf{fMRI-to-T1w co-registration:} A 3$\times$3 montage including axial, sagittal, and coronal views. The background image is the co-registered mean BOLD/EPI image. T1w-derived contours are overlaid as red outlines, including the outer brain boundary and internal white-matter boundaries (Figure~\ref{fig:fmri_coreg_to_t1w_qc_vis}).
\end{itemize}

\begin{figure}[ht]
    \centering
    \includegraphics[width=0.72\textwidth]{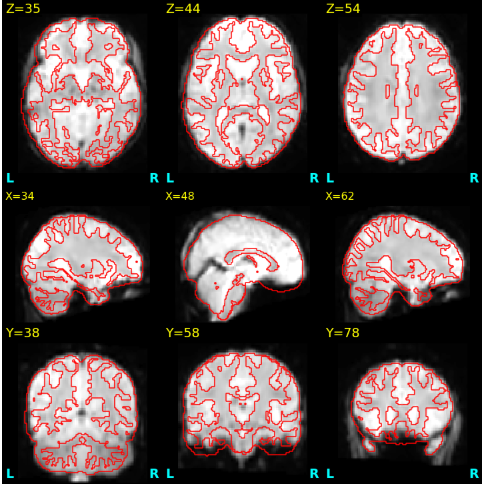}
    \caption{Example visualization used for fMRI-to-MNI normalization QC. The red contours correspond to MNI template boundaries overlaid on the normalized mean BOLD image.}
    \label{fig:fmri_norm_to_mni_qc_vis}
\end{figure}

\begin{itemize}
    \item \textbf{fMRI-to-MNI normalization:} A 3$\times$3 montage including axial, sagittal, and coronal views. The background image is the normalized mean BOLD/EPI image. The MNI template contours are overlaid as red outlines, including the outer brain boundary and internal white-matter boundaries (Figure~\ref{fig:fmri_norm_to_mni_qc_vis}).
\end{itemize}

\newpage

\begin{table*}[t]
\centering
\footnotesize
\caption{Quality control primitives used in NEXUS.}
\label{tab:qc_primitive_lib}
\renewcommand{\arraystretch}{1.15}
\setlength{\tabcolsep}{5pt}
\begin{tabularx}{\textwidth}{
>{\raggedright\arraybackslash}p{0.45\textwidth}|
>{\raggedright\arraybackslash}X}
\hline
\textbf{Primitive Name} & \textbf{Primitive Description} \\
\hline
\texttt{\small run\_mriqc} &
Runs MRIQC on raw BIDS-formatted neuroimaging data to compute image quality metrics and generate subject-level visual QC reports before preprocessing. \\
\hline

\texttt{\small get\_mriqc\_group\_level\_tsv\_and\_outliers} &
Reads group-level MRIQC results and identifies IQM outliers across subjects using metric-based cohort screening. \\
\hline

\texttt{\small read\_mriqc\_metric} &
Retrieves subject-level MRIQC image quality metrics for a specified modality, optionally restricted to selected metric keys. \\
\hline

\texttt{\small mriqc\_visual\_inspection\_based\_qc} &
Performs visual QC on MRIQC-generated subject reports for selected raw-data visualization elements. \\
\hline

\texttt{\small smri\_skull\_stripping\_qc} &
Performs post-processing QC for structural MRI skull stripping using either brain-volume metrics or visual inspection. \\
\hline

\texttt{\small smri\_tissue\_segmentation\_qc} &
Performs post-processing QC for structural MRI tissue segmentation using either tissue-volume metrics or visual inspection. \\
\hline

\texttt{\small smri\_to\_mni\_normalization\_qc} &
Performs post-processing QC for structural MRI normalization to MNI space using similarity metrics or visual inspection. \\
\hline

\texttt{\small fmri\_coregister\_to\_anat\_qc} &
Performs post-processing QC for fMRI-to-anatomical co-registration using registration quality metrics or visual inspection. \\
\hline

\texttt{\small fmri\_normalized\_to\_mni\_qc} &
Performs post-processing QC for fMRI normalization to MNI space using similarity metrics or visual inspection. \\
\hline
\end{tabularx}
\end{table*}


\begin{table*}[t]
\centering
\footnotesize
\caption{Processing primitives used in NEXUS.}
\label{tab:processing_primitive_lib}
\renewcommand{\arraystretch}{1.15}
\setlength{\tabcolsep}{5pt}
\begin{tabularx}{\textwidth}{
>{\raggedright\arraybackslash}p{0.39\textwidth}|
>{\raggedright\arraybackslash}X}
\hline
\textbf{Primitive Name} & \textbf{Primitive Description} \\
\hline
\texttt{\small fsl\_bet\_t1w} &
Performs structural MRI skull stripping using FSL BET, producing a brain-extracted T1w image and brain mask. \\
\hline

\texttt{\small fsl\_fast} &
Performs structural MRI tissue segmentation using FSL FAST, producing CSF/GM/WM maps and a hard segmentation for QC. \\
\hline

\texttt{\small fsl\_normalize\_t1w\_to\_mni} &
Normalizes a skull-stripped T1w image to MNI space using FSL affine and nonlinear registration. \\
\hline

\texttt{\small fsl\_warp\_gm\_tissue\_seg\_to\_mni} &
Warps the gray-matter probability map from native space to MNI space using FSL. \\
\hline

\texttt{\small fsl\_skullstrip\_fmri} &
Performs brain extraction on 4D fMRI data using FSL BET, producing a skull-stripped image and brain mask. \\
\hline

\texttt{\small fsl\_slicetimer} &
Performs slice-timing correction on fMRI data using FSL Slicetimer. \\
\hline

\texttt{\small fsl\_motion\_correct} &
Performs motion correction on fMRI data using FSL MCFLIRT, producing corrected data, motion parameters, and a QC mask. \\
\hline

\texttt{\small fsl\_coregister\_func\_to\_anat} &
Co-registers fMRI data to anatomical T1w space using FSL and produces QC-relevant derivatives. \\
\hline

\texttt{\small fsl\_normalize\_func\_to\_mni} &
Normalizes fMRI data to MNI space using FSL after functional-to-anatomical registration. \\
\hline

\texttt{\small fsl\_spatial\_smooth} &
Applies spatial smoothing to fMRI data using FSL. \\
\hline

\texttt{\small fsl\_temporal\_filter} &
Applies temporal band-pass filtering to fMRI data using FSL. \\
\hline

\texttt{\small fmriprep\_pipe} &
Runs fMRIPrep for end-to-end preprocessing of a subject’s structural and functional MRI data, producing analysis-ready derivatives and QC-related outputs. \\
\hline

\texttt{\small ants\_bias\_field\_correction} &
Performs N4 bias-field correction on structural MRI using ANTs. \\
\hline

\texttt{\small ants\_skull\_strip} &
Performs structural MRI skull stripping using ANTs brain extraction. \\
\hline

\texttt{\small ants\_tissue\_segmentation} &
Performs structural MRI tissue segmentation using ANTs Atropos, producing tissue maps and a hard segmentation for QC. \\
\hline

\texttt{\small ants\_normalize\_to\_mni\_template} &
Normalizes structural MRI to MNI template space using ANTs registration. \\
\hline

\texttt{\small ants\_warp\_gm\_tissue\_seg\_to\_mni} &
Warps the gray-matter probability map from native space to MNI space using ANTs. \\
\hline

\texttt{\small ants\_cortical\_thickness\_estimate} &
Warps the gray-matter probability map from native space to MNI space using ANTs. \\
\hline

\texttt{\small afni\_t1w\_skull\_strip} &
Perform cortical thickness estimation using ANTs antsCorticalThickness.sh. \\
\hline

\texttt{\small afni\_tissue\_segmentation} &
Performs structural MRI tissue segmentation using AFNI 3dSeg, producing GM probability and hard segmentation maps. \\
\hline

\texttt{\small afni\_normalize\_t1w\_to\_mni} &
Normalizes structural MRI to MNI template space using AFNI. \\
\hline

\texttt{\small afni\_warp\_gm\_tissue\_seg\_to\_mni} &
Warps the gray-matter probability map from native space to MNI space using AFNI. \\
\hline

\texttt{\small afni\_fmri\_slice\_timing\_correction} &
Performs slice-timing correction on fMRI data using AFNI 3dTshift. \\
\hline

\texttt{\small afni\_fmri\_motion\_correction} &
Performs motion correction on fMRI data using AFNI 3dvolreg, producing corrected data, motion parameters, and a QC mask. \\
\hline

\texttt{\small afni\_coregister\_fmri\_to\_anat} &
Co-registers fMRI data to anatomical space using AFNI and produces QC-related anatomical and functional derivatives. \\
\hline

\texttt{\small afni\_normalize\_fmri\_to\_mni} &
Normalizes fMRI data to MNI space using AFNI after T1w normalization, producing both MNI-space outputs and QC-related intermediate derivatives. \\
\hline

\texttt{\small afni\_spatial\_smoothing} &
Applies spatial smoothing to fMRI data using AFNI. \\
\hline

\texttt{\small afni\_temporal\_filter\_fmri} &
Performs temporal filtering and nuisance regression on fMRI data using AFNI. \\
\hline
\end{tabularx}
\end{table*}


\begin{table*}[t]
\centering
\footnotesize
\caption{Downstream analysis primitives used in NEXUS.}
\label{tab:downstream_primitive_lib}
\renewcommand{\arraystretch}{1.15}
\setlength{\tabcolsep}{5pt}
\begin{tabularx}{\textwidth}{
>{\raggedright\arraybackslash}p{0.46\textwidth}|
>{\raggedright\arraybackslash}X}
\hline
\textbf{Primitive Name} & \textbf{Primitive Description} \\
\hline
\texttt{\small construct\_functional\_brain\_connectivity} &
Constructs a subject-level functional brain connectivity matrix from preprocessed fMRI data and a chosen parcellation atlas. \\
\hline

\texttt{\small BrainNetworkTransformer} &
A state-of-the-art graph transformer model for functional brain connectome analysis, supporting classification or regression tasks. \\
\hline

\texttt{\small sparsify\_connectivity\_matrices} &
Sparsifies functional connectivity matrices by retaining a subset of edges, typically for graph-based downstream modeling. \\
\hline

\texttt{\small NeuroGraph} &
A state-of-the-art graph neural network model for functional brain connectome analysis with residual connections, supporting classification or regression. \\
\hline
\end{tabularx}
\end{table*}


\clearpage

\end{document}